\documentclass[letterpaper]{article} %
\usepackage{aaai24}  %
\usepackage{times}  %
\usepackage{helvet}  %
\usepackage{courier}  %
\usepackage[hyphens]{url}  %
\usepackage{graphicx} %
\usepackage{natbib}  %
\usepackage{caption} %
\usepackage{algorithm}

\usepackage{newfloat}
\usepackage{listings}
\DeclareCaptionStyle{ruled}{labelfont=normalfont,labelsep=colon,strut=off} %
\floatstyle{ruled}
\newfloat{listing}{tb}{lst}{}
\floatname{listing}{Listing}
\usepackage{amsmath,amssymb}

\usepackage{colortbl}
\usepackage[dvipsnames,table,xcdraw]{xcolor}
\usepackage{array}
\usepackage{nccmath}
\usepackage{booktabs}
\usepackage{tabularx}
\usepackage{algpseudocode}
\usepackage{todonotes}
\usepackage{rotating}
\usepackage{multirow}
\usepackage{arydshln}

\definecolor{mygreen}{RGB}{112,173,71}

\usepackage{xspace}
\makeatletter
\DeclareRobustCommand\onedot{\futurelet\@let@token\@onedot}
\def\@onedot{\ifx\@let@token.\else.\null\fi\xspace}
\def\eg{\emph{e.g}\onedot} 
\def\ie{\emph{i.e}\onedot} 
 
\def\etc{\emph{etc}\onedot}

\makeatother

\definecolor{MyGreen}{RGB}{84,130,53}
\definecolor{MyOrange}{RGB}{197,90,17}
\definecolor{MyGray}{RGB}{127,127,127}

\title{\textit{HOP} to the Next Tasks and Domains for Continual Learning in NLP}
\author{
    Umberto Michieli, Mete Ozay
}
\affiliations{
    Samsung Research UK\\
    \{u.michieli, m.ozay\}@samsung.com

}

\usepackage{bibentry}

\begin{document}

\maketitle

\begin{abstract}
Continual Learning (CL) aims to learn a sequence of problems (\ie, tasks and domains) by transferring knowledge acquired on previous problems, whilst avoiding forgetting of past ones.
Different from previous approaches which focused on CL for one NLP task or domain in a specific use-case, in this paper, we address a more general CL setting to learn from a sequence of problems in a unique framework.
Our method, HOP, permits to \textit{hop} across tasks and domains by addressing the CL problem along three directions: (i) we employ a set of adapters to generalize a large pre-trained model to unseen problems, (ii) we compute high-order moments over the distribution of embedded representations to distinguish independent and correlated statistics across different tasks and domains, (iii) we process this enriched information with auxiliary heads specialized for each end problem.
Extensive experimental campaign on 4 NLP applications, 5 benchmarks and 2 CL setups demonstrates the effectiveness of our HOP.

\end{abstract}

\section{Introduction}

Current practice to obtain a deep learning model to perform a specific assignment is to train the model on a specific dataset for that particular assignment \cite{chen2018lifelong}.
However, this paradigm has several inherent limitations.
For example, as models get larger, training from scratch requires a larger amount of expensive labeled data and computation time, which can be reduced by knowledge transfer (KT) from a pre-trained model on a base domain. 
This problem is generally addressed via adapting (\eg, through fine-tuning, FT) large pre-trained language models (\eg, BERT, \citealp{delvin2019bert}) to various downstream NLP applications, 
such as Text Classification (TC, \citealp{wang2023effective,wu2023continual,ke2021adapting,hu2021continual}), Natural Language Inference (NLI, \citealp{pfeiffer2020mad}), Document \cite{ke2020continual} or Aspect \cite{ke2021adapting,zhou2021adaptive} Sentiment Classification (DSC and ASC).
In our work, we employ the Adapter-BERT \cite{houlsby2019parameter} model.
FT a large pre-trained model reaches state-of-the-art results on NLP benchmarks with a \textit{static} distribution. 
However, if a stream of problems\footnote{For the sake of clarity, we refer to \textit{problems} as either \textit{tasks} or \textit{domains} experienced by the CL method over time.} are presented sequentially, the na\"ively FT model faces catastrophic forgetting (CF, \citealp{mccloskey1989catastrophic}) of previous knowledge due to the non-stationary data distribution, and cannot make use of past knowledge to improve capability on subsequent problems (forward KT, \citealp{lopez2017gradient}) or vice-versa (backward KT). In particular, high CF and low KT hinder performance of CL for NLP, as several NLP applications share similar knowledge that can be exploited to achieve higher accuracy on future/previous problems, without accuracy degradation. Indeed, ideally, learning a sequence of problems should allow multiple problems to support each other via KT \cite{rusu2016progressive,ke2020continualnips}.
To address such issues towards more versatile NLP models, we focus on CL to tune a pre-trained Adapter-BERT for a stream of problems. 

In CL, a model learns a sequence of problems incrementally. 
After each 
incremental learning (IL) stage is completed, its training data is typically discarded \cite{chen2018lifelong}. 
Three main families of CL setups can be identified \cite{van2019three}: namely, Task-IL (TIL), Domain-IL (DIL), and Class-IL (CIL). 
TIL builds one model for each task (\eg, to classify sentiment in products' reviews). At test time, task identifier specifies the proper model for each input sample. This could significantly increase the number of parameters; however, in our case, most of the parameters are shared across problems and only a few parameters are problem-specific.
DIL is similar to TIL, however, builds a single head for each domain as classes are shared across domains. In DIL, no identifier is required at test time and subsequent problems present data from different domains (\eg, reviews from online commerce, or from movie critique, \etc).
In CIL, non-overlapping classes are learned progressively.
Opposed from traditional CL approaches used in Computer Vision (CV), most of the NLP problems are formulated as either TIL or DIL \cite{biesialska2020continual,ke2022continual,ke2021classic,sun2019lamol} and, to the best of our knowledge, no prior work has addressed them both jointly. %

Differently from concurrent CL NLP approaches, in this paper, we evaluate models on both TIL and DIL in a unified framework which employs  parameter-efficient transfer learning strategies to adapt the models to each end problem: (i) as in current state-of-the-art approaches \cite{ke2021adapting,ke2020continualnips}, we use Adapter-BERT with a separate set of adapters tuned for each problem; (ii) we  discard the \texttt{[CLS]} token, which we show being counterproductive for CL, and rather compute high-order statistical measures over the distribution of extracted features (\ie, tokens representing text embeddings); (iii) we use an MLP head specialized for each problem to process and combine such information.
With a slight abuse of notation, we call our method HOP, from High-Order Pooling, that is the most distinctive component of our method. %

Our HOP extracts multiple cues from the limited samples drawn from  non-stationary distributions while preserving previous knowledge. HOP accurately models the variable distribution of problems since input-level distribution shift is reflected into feature-level distribution shift. Indeed, we show that this variation is not properly captured by the single \texttt{[CLS]} token. 
We present an extensive validation on 2 CL setups (DIL and TIL) and 5 benchmarks outperforming current state-of-the-art on accuracy, KT, CF, and runtime.

\section{Related Work}
\label{sec:related}

For a wide survey of the current state-of-the-art, we refer to recent CL reviews \cite{de2021continual,lesort2020continual,michieli2022275,biesialska2020continual,ke2022continual}.

\noindent\textbf{Traditional CL methods} 
focused on image classification and can be grouped according to the proposed technique. 
(1) Regularization-based methods are generally based on knowledge distillation  \cite{li2017learning,jung2016less,michieli2019incremental} %
or on importance score for each parameter to compute a penalty term in the optimization to reduce weight deviation while learning new problems \cite{kirkpatrick2017overcoming,zeng2019continual,nguyen2018variational,zenke2017continual,ahn2019uncertainty}.
(2) Parameter-isolation approaches dedicate a set of parameters to each problem to reduce forgetting when learning subsequent problems. Parameters can be either masked out \cite{serra2018overcoming,mallya2018packnet,mallya2018piggyback,wang2023isolation}, frozen \cite{rusu2016progressive,xu2018reinforced,michieli2021knowledge}, or new branches are grown over time \cite{rusu2016progressive,xu2018reinforced}.
(3) Replay-based methods either retain an exemplar set of previously seen data \cite{rebuffi2017icarl,isele2018selective,lopez2017gradient,chaudhry2018efficient} %
or generated pseudo-samples \cite{shin2017continual,maracani2021recall} to reduce CF and promote KT. %

\noindent\textbf{CL in NLP} is in rapid expansion due to its great importance.
Early works tackled lifelong learning (with no knowledge preservation, hence no CF) for sentiment analysis \cite{carlson2010toward,silver2013lifelong,ruvolo2013ella,chen2015lifelong,wang2019forward,qin2020using,wang2018lifelong}. %
Recent works have dealt with CF in many applications: sentiment analysis \cite{lv2019sentiment,ke2021classic,ke2021adapting}, dialogue systems \cite{shen2019progressive,madotto2020continual,qian2021student,chien2021continuous}, language modeling \cite{sun2019lamol,chuang2020lifelong} and learning \cite{li2019compositional}, cross-lingual modeling \cite{liu2020exploring}, sentence embedding \cite{liu2019continual}, machine translation \cite{khayrallah2018regularized,zhan2021meta}, question answering \cite{greco2019psycholinguistics}, named entity recognition \cite{monaikul2021continual}.

Most of the previous literature focuses on the simpler TIL setup. SRK \cite{lv2019sentiment} and KAN \cite{ke2020continual} tackled DSC via recurrent architectures. They are mainly conceived for KT, hence they suffer from CF and cannot be easily extended to BERT.
B-CL \cite{ke2021adapting} is the first CL framework for ASC: it employs Adapter-BERT and is based on capsule network and dynamic routing, bringing only limited KT.
CAT \cite{ke2020continualnips} works on mixed sequences of similar and dissimilar problems, and can transfer knowledge among similar problems.
Snapshot \cite{wang2023effective} regularizes training with adapters learned over previous problems.
Parallel to CL, AdapterFusion \cite{pfeiffer2021adapterfusion} uses a two-stage method to learn the adapters of Adapter-BERT to improve multi-task learning, hence it has no CF. CTR \cite{ke2021achieving} extends the adapters concept to the idea of CL plugins to adapt BERT to each problem, and it is the state-of-the-art in TIL.

Recently, DIL has gained attention.
LAMOL \cite{sun2019lamol} uses a language model (\ie, GPT-2 \citealp{radford2019language}) to solve sequential problems and to generate training pseudo-samples against CF;
CLASSIC \cite{ke2021classic} uses contrastive learning to promote KT. %

\textbf{Pooling in NLP} has been recently studied to improve accuracy \cite{wu2020attentive,acs2021subword,zhao2022augment}. In particular, \citealp{wu2020attentive} propose an attentive pooling scheme with learnable norm to extract accurate text representations in different problems, motivated by 3 observations: (i) different contexts have different informativeness for learning text representations (\eg, they might be important to determine sentiment polarity, however, probably less relevant for TC); (ii)  different problems have different characteristics; (iii) popular pooling methods (\eg, MAX or AVG) may over-emphasize some concepts and disregard other useful contextual information. To summarize, some problem specific words or sentences contain information regarding output class in various ways.
Our work is motivated by these results.
However, such pooling schemes cannot be applied to CL.
To cope with this, our HOP computes multiple statistical moments from the encoded text to capture evolution of different statistics of the input.

\section{Problem Formulation}

CL learns a sequence of problems ${t\in\{1,\dots,T\}}$. 
Each problem $t$ has its test data $\hat{\mathcal{S}}_t$ and training data ${{\mathcal{S}}_t = \{ (x_{t}^k, y_{t}^k) \}_{k=1}^{N_t0}}$, where ${x_{t}^k\in\mathcal{X}_t}$  is a training sample with label $y_{t}^k\in\mathcal{Y}_t$ (\ie, supervised problems). Then, the CL goal is to minimize the empirical loss $\mathcal{L}$ over all seen problems. At problem $T$, we aim at training models $f_t, \forall t$, 
parameterized by $\theta$ (\ie, $\hat{y}_t^k = f_t(x_{t}^k; \theta)$), which minimize the loss
\begin{equation}
    \label{eq:loss}
   \sum_{t=1}^{T} l_t,
    \quad
    \mathrm{with} \ \
    l_t \!=\! %
    \frac{1}{N_{t}}
    \sum_{k=1}^{N_{t}}
    \mathcal{L}(\hat{y}_t^k, y_{t}^k )
    .
\end{equation}
However, Eq.~\eqref{eq:loss} cannot be minimized since no (in case of replay-free CL methods) or limited (in case of replay CL methods) access to previous data is guaranteed. In the most challenging replay-free setup, we can minimize the empirical loss on the current problem $T$ only, \ie, $l_T$.

Therefore, CL methods try to approximate Eq.~\eqref{eq:loss} in different ways (\eg, via regularization, replay, \etc). Instead, we extract high-order statistics from the input dataset and we process this additional information via an auxiliary problem-specific multi-layer perceptron (MLP) to adapt the current model to the current problem.

Depending on the properties of  $(\mathcal{X}_t, \mathcal{Y}_t)$ $\forall t$, we can identify TIL, DIL and CIL. %
In CIL, models are progressively trained with new classes, and it has been less attractive for NLP applications as the number of classes is generally determined \textit{a priori} \cite{ke2021classic}. Therefore, we address both TIL and DIL in a unified framework of CL for NLP.

\begin{figure}
\centering
    \includegraphics[trim=0cm 14.8cm 19.3cm 0cm, clip, width=\linewidth]{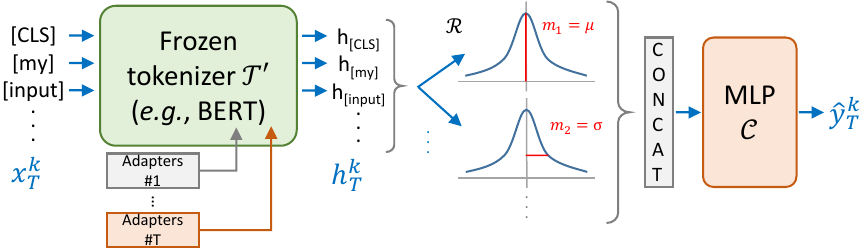}
    \caption{The proposed HOP framework. During the incremental step $T$, only orange modules are trained, while gray and green modules are frozen.}
    \label{fig:architecture}
 \end{figure}
\begin{figure}%
    \centering
    \includegraphics[width=\linewidth, trim=0cm 14.1cm 22.4cm 0cm, clip]{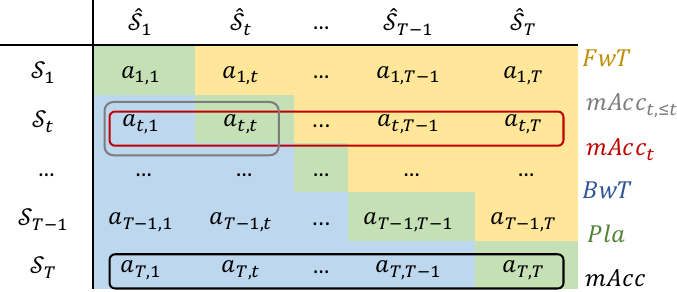}
    \caption{Accuracy matrix showing the main CL metrics used in this work. $\hat{\mathcal{S}}_t$ and ${\mathcal{S}}_t$ are the testing and training datasets at step $t$.}
    \label{fig:CL_metrics}
\end{figure}

\section{Our Proposed Framework: HOP}

Our framework employs parameter-efficient learning strategies to adapt models to each end problem, namely: (i) adapter modules, (ii) computation of high-order moments, and (iii) a specialized MLP for each end problem.
In the experimental validation, we apply HOP to BERT-based models due to its superior performance in NLP, however, it can be seamlessly applied to other architectures as well.

\noindent (i) Following recent CL approaches for NLP \cite{ke2021adapting,ke2020continualnips}, our best architecture relies on Adapter-BERT, with a separate set of adapters tuned for each problem. 
Adding adapters to BERT is a highly parameter-efficient transfer learning paradigm: in CL, this means that subsequent problems have separate adapters (which are small in size). %
An adapter layer is a tunable 2-layer fully-connected network, which adapts the pre-trained model to the end problem at hand. 
In this way, there is no need for a separate BERT model fine-tuned on each problem, which is extremely parameter-inefficient if many problems are learned in sequence.

\noindent  (ii) HOP computes high-order statistical moments from the distribution of extracted tokens to capture most of the information from the input sequence. 
Current approaches \cite{ke2021adapting,ke2021achieving} design their CL systems relying on the \texttt{[CLS]} token embeddings. 
More formally, we consider $f_t$ composed of a tokenizer $\mathcal{T}$ and a classifier $\mathcal{C}$ by $f_t = \mathcal{C} \circ \mathcal{T}$ to recognize $N_C$ classes ($\circ$ represents function composition).
Traditionally, $\mathcal{T}$ includes a reduction function $\mathcal{R}$ as the last layer to summarize the whole input sequence into one element. Therefore, we can write $\mathcal{T}=\mathcal{R} \circ \mathcal{T}'$, with $\mathcal{T}'$ being the tokenizer without the final reduction function. $\mathcal{R}$ is often identified by the \texttt{[CLS]} token or by AVG pooling.
However, different problems usually have different peculiar patterns in the input samples, and the output should be an explicit function of the whole, non-reduced, embedding sequence.
Therefore, we compute high-order \textit{central} moments from the input sequence and concatenate ($\mathrm{concat}$) them.
We define the reduction function by
\begin{equation}
    \mathcal{R} = \mathrm{concat}
    (m_1, m_2, \dots, m_p)
    ,
\end{equation}
where $p$ is the order of considered moments, that is, $m_1$ is the first moment (\ie, AVG), $m_2$ is the second moment (\ie, the variance), \etc \cite{papoulis2002probability,michieli2023online1,michieli2023online2,michieli2024object}.
Such moments are computed over the distribution of tokens identified by the unreduced tokenizer $\mathcal{T}': x_t^k \mapsto {h_{t,d}^k}$ where $d$ denotes the dimension of the embedded sequence and each $h_{t,d}^k \in \mathbb{R}^{Q}$ with $Q$ channels. \\
This step may recall the statistical moments of tokens identified by standard N-gram models \cite{cortes2004distribution}.

\noindent (iii) We process and combine embeddings computed by $\mathcal{T}$ with an auxiliary MLP head specialized for each problem. We replace the usual linear layer constituting $\mathcal{C}$ with an MLP.
The MLP head increases the adaptation capacity to process the high order information while being highly parameter efficient.
We design the MLP as a 2-layer network consisting of $p \cdot Q$ and $N_C$ neurons at each layer, respectively. 

Overall, our HOP can extract richer information from the samples drawn from the non-stationary input sequence distributions while preserving previous knowledge. Therefore, our method can \textit{hop} across the distributions of subsequent tasks and domains, since input-level distribution shift is reflected into a feature-level distribution shift via the embedding tokenizer. Our framework is applicable to both TIL and DIL setups: in TIL, one head per task adapts models to each of the tasks separately; in DIL, one single head is incrementally adapted to the varying domains.

HOP alleviates CF thanks to (i) frozen shared backbone to extract shared rich features, and (ii) adapters tuned for each problem. It promotes KT via (i) initialization of adapters to the last achieved ones, therefore bringing onward the previous information; (ii) modeling tailed distributions of tokens encountered in subsequent CL problems with different moments. Finally, MLP heads improve plasticity for new tasks.

An overview of our framework is given in Figure~\ref{fig:architecture}, where we depict the modular property of HOP, so it can be plugged on top of other CL methods. Compared to competing methods, HOP only brings a minimal computation and complexity footprint. Detailed analyses are provided in Sec.~\ref{sec:results}. 

The main novelty of our HOP are: (i) we are the first to address tailed distribution of features and its relation to CL; (ii) we revisit high-order feature statistics (inspired from N-grams co-occurrence statistics) to deep NLP architectures; (iii) we exploit them in CL for the first time; (iv) we use adapters for CL extending concurrent works to jointly tackle TIL/DIL; (v) we include specialized MLP heads to improve plasticity in learning new concepts.
Overall, we propose a simple yet effective CL baseline that outperforms complex architectures in all experiments, while requiring significantly lower training time. Additionally, it can be seamlessly applied on top of competitors.

\section{Experimental Setup}

\textbf{Architectures.} We evaluate HOP on 3 BERT-based architectures \cite{delvin2019bert}.
Previous works \cite{ke2021achieving,ke2021adapting} have shown that na\"{i}vely fine-tuning BERT increases CF, hence we focus on more versatile ways to learn new concepts limiting CF. 
{\textit{BF + Lin}} consists in a Frozen BERT (BF) with a trainable linear layer on top.
\textit{BF + CNN} consists in a BF with a trainable CNN TC network \cite{kim2014convolutional} on top.
\textit{Adapter-BERT} trains only the adapter blocks built of 2 linear layers each with $2000$ neurons.

\noindent\textbf{Datasets.} We consider 4 applications, 
unifying the setups proposed by \cite{ke2021classic,ke2021achieving,asghar2019progressive}. 
(1) ASC classifies a review sentence on either \textit{positive}, \textit{negative} or \textit{neutral} aspect-level sentiments. We use 19 datasets (\ie, reviews of 19 products) taken from four sources: 5 products from HL5Domains \cite{hu2004mining}, 3 products from Liu3Domains \cite{liu2015automated}, 9 products from Ding9Domains \cite{ding2008holistic}, and 2 products from SemEval14 Task 4 \cite{pontiki2014semeval}. We applied the same data filtering of previous works for fair comparison \cite{ke2021achieving}.
(2)  DSC classifies product reviews into either \textit{positive} or \textit{negative} opinion classes, using TC formulation of \citealp{delvin2019bert}. We use 10 DSC datasets (\ie, reviews of 10 products, \citealp{ke2020continual}). We consider both a small training version of 100 positive and 100 negative reviews per problem, and the full training version of 2500 positive and 2500 negative reviews per problem. Validation and test sets are fixed, and each consists of 250 reviews per each class. The first experiment is arguably more useful in practice because labeling a large number of examples is costly, thus, ablation is carried out on this split.
(3) TC classifies text into 20 classes using 20News data \cite{lang1995newsweeder}. We split documents into 10 problems with 2 classes per problem (in DIL, $N_C$ is supposed known \textit{a priori}).
Classes are variegate and share little knowledge, hence forgetting is the main issue.
(4) We target NLI for sentence understanding using the MultiNLI dataset \cite{williams2018broad} which is one of the largest corpus of its kind. Sentences are classified into: \textit{entailment}, \textit{neutral} and \textit{contradiction}. We split data in 5 problems, each belonging to a specific domain (\textit{fiction}, \textit{telephone}, \etc, \citealp{asghar2019progressive}). %

\noindent\textbf{Baselines.} As the first baseline, we consider a separate model learned for each problem independently, which we call SDL (standalone). No KT or CF occurs here. 
Second, we compare against FT which simply optimizes the model over the sequence of problems.
Third, we examine 13 CL competitors. Among them, some approaches have been proposed for CL in NLP and have been briefly described in Sec.~\ref{sec:related}.
Additionally, we 
adapted CL methods proposed in CV for our NLP applications. For TIL-based works: UCL \cite{ahn2019uncertainty} proposes uncertainty-based regularization via a Bayesian online learning framework; HAT \cite{serra2018overcoming} focuses on employing embeddings preserving information of previous problems while learning new ones.
CIL approaches can be adapted to TIL by training the head of the specific problem and considering predictions of the specific head during testing. Among them, we used (i) regularization-based methods, \eg,
EWC \cite{kirkpatrick2017overcoming},
OWM \cite{zeng2019continual}, and
L2 \cite{kirkpatrick2017overcoming}, and (ii)
replay-based methods, \eg,
the efficient A-GEM \cite{chaudhry2018efficient},
and DER++ \cite{buzzega2020dark} for pseudo replay.
We note that a single upper bound for accuracy 
does not exist, since different methods have a different number of parameters, as shown in Table~\ref{tab:cf_kt_params_time_metrics}.

\noindent\textbf{Hyperparameters.}
We employ the same scenarios as current state-of-the-art approaches \cite{ke2021achieving}.
We follow the CL evaluation of \cite{de2021continual}: 
after training on one problem is completed, the respective training data is no longer accessible; 
all hyperparameters are chosen according to the performance on the validation set; 
after all problems are learned, testing is carried out on the test set.
We report results averaged over $5$ random seeds (\ie, different ordering of problems) and we report the mean (the standard deviation is negligible - lower than $0.1$ in all cases).
All baseline approaches use the \texttt{[CLS]} token as the reduction function of the tokenizer. We show that this is a major limitation, concurrently hinted also by \cite{mirzadeh2022architecture}. However, \cite{mirzadeh2022architecture} address the issue only marginally, while we propose a simple and effective framework to overcome it.
The only hyperparameter specific to our framework is $p$, which is set according to the best validation results. %
Empirically, $p=3$ provides the best results and represents a good compromise with additional computational complexity. 

\begin{table*}[!htb]
\setlength{\tabcolsep}{1.3pt}
  \centering
  \caption{$mAcc$ and $MF1$ over 5 benchmark datasets on both TIL and DIL setups. We evaluate 3 network architectures based on BERT and 14 baselines. Results are color-coded according to the column-wise value and best results are in \textbf{bold}.}
  \resizebox{0.9\linewidth}{!}{%
    \begin{tabular}{clcccccccccccccccccccc}
    \toprule
    \multirow{3}[0]{*}{} & \multicolumn{1}{c}{\multirow{3}[0]{*}{\textbf{CL Method}}} & \multicolumn{4}{c}{\textbf{ASC}} & \multicolumn{4}{c}{\textbf{DSC (small)}} & \multicolumn{4}{c}{\textbf{DSC (full)}} & \multicolumn{4}{c}{\textbf{20News}} & \multicolumn{4}{c}{\textbf{NLI}} \\
    \cmidrule(lr){3-6} \cmidrule(lr){7-10} \cmidrule(lr){11-14} \cmidrule(lr){15-18} \cmidrule(lr){19-22}
    
          &       & \multicolumn{2}{c}{\textbf{TIL}} & \multicolumn{2}{c}{\textbf{DIL}} & \multicolumn{2}{c}{\textbf{TIL}} & \multicolumn{2}{c}{\textbf{DIL}} & \multicolumn{2}{c}{\textbf{TIL}} & \multicolumn{2}{c}{\textbf{DIL}} & \multicolumn{2}{c}{\textbf{TIL}} & \multicolumn{2}{c}{\textbf{DIL}} & \multicolumn{2}{c}{\textbf{TIL}} & \multicolumn{2}{c}{\textbf{DIL}} \\ \cmidrule(lr){3-4}\cmidrule(lr){5-6}\cmidrule(lr){7-8}\cmidrule(lr){9-10}\cmidrule(lr){11-12}\cmidrule(lr){13-14}\cmidrule(lr){15-16}\cmidrule(lr){17-18}\cmidrule(lr){19-20}\cmidrule(lr){21-22}
          
          &       & $\mathbf{mAcc}$ & $\mathbf{MF1}$ & $\mathbf{mAcc}$  & $\mathbf{MF1}$ & $\mathbf{mAcc}$  & $\mathbf{MF1}$ & $\mathbf{mAcc}$  & $\mathbf{MF1}$ & $\mathbf{mAcc}$  & $\mathbf{MF1}$ & $\mathbf{mAcc}$  & $\mathbf{MF1}$ & $\mathbf{mAcc}$  & $\mathbf{MF1}$ & $\mathbf{mAcc}$  & $\mathbf{MF1}$ & $\mathbf{mAcc}$  & $\mathbf{MF1}$ & $\mathbf{mAcc}$  & $\mathbf{MF1}$ \\
          \midrule
    
    \multirow{3}[0]{*}{\begin{sideways}\textbf{BF+Lin}\end{sideways}} & SDL   & 56.32 & 27.8  & 58.48 & 35.21 & 48.97 & 42.21 & 57.87 & 56.97 & 82.32 & 80.11 & 77.45 & 76.94 & 95.31 & 95.30  & 54.96 & 54.31 & 48.69 & 43.78 & 43.24 & 41.55 \\\cdashline{2-22}
          & FT    & 61.18 & 32.00 & 68.42 & 49.97 & 60.41 & 52.36 & 66.12 & 65.06 & 70.53 & 63.62 & 75.07 & 74.19 & 66.30 & 65.59 & 55.85 & 54.88 & 42.53 & 37.09 & 41.08 & 38.17 \\
          & HOP (ours) & \textbf{86.38} & \textbf{78.64} & \textbf{87.54} & \textbf{81.79} & \textbf{78.48} & \textbf{75.31} & \textbf{81.96} & \textbf{80.93} & \textbf{84.59} & \textbf{83.18} & \textbf{85.63} & \textbf{84.65} & \textbf{96.45} & \textbf{96.45} & \textbf{68.45} & \textbf{66.16} & \textbf{61.05} & \textbf{60.88} & \textbf{59.24} & \textbf{58.82} \\
          \midrule
    
    \multirow{13}[0]{*}{\begin{sideways}\textbf{Bert (frozen) + CNN}\end{sideways}} & SDL   & 78.14 & 58.13 & 78.14 & 58.13 & 73.88 & 67.97 & 73.88 & 67.97 & 85.34 & 80.17 & 85.34 & 80.17 & 96.49 & 96.48 & 96.49 & 96.48 & 68.21 & 67.49 & 68.21 & 67.49 \\\cdashline{2-22}
          & FT    & \cellcolor[rgb]{ 1,  .922,  .518}85.51 & \cellcolor[rgb]{ .996,  .914,  .514}76.64 & \cellcolor[rgb]{ .8,  .867,  .51}86.85 & \cellcolor[rgb]{ .851,  .878,  .51}78.73 & \cellcolor[rgb]{ .459,  .769,  .49}83.12 & \cellcolor[rgb]{ .541,  .792,  .494}79.23 & \cellcolor[rgb]{ .4,  .749,  .486}85.66 & \cellcolor[rgb]{ .404,  .753,  .486}84.87 & \cellcolor[rgb]{ .973,  .412,  .42}61.88 & \cellcolor[rgb]{ .973,  .412,  .42}45.79 & \cellcolor[rgb]{ .996,  .898,  .51}85.54 & \cellcolor[rgb]{ .996,  .898,  .51}84.59 & \cellcolor[rgb]{ .996,  .906,  .514}83.28 & \cellcolor[rgb]{ .996,  .914,  .514}81.81 & \cellcolor[rgb]{ .839,  .878,  .51}64.45 & \cellcolor[rgb]{ .871,  .886,  .514}58.68 & \cellcolor[rgb]{ .957,  .91,  .518}71.98 & \cellcolor[rgb]{ .976,  .918,  .518}71.70 & \cellcolor[rgb]{ .996,  .914,  .514}71.70 & \cellcolor[rgb]{ .996,  .91,  .514}71.47 \\
          & L2    & \cellcolor[rgb]{ .973,  .412,  .42}56.04 & \cellcolor[rgb]{ .973,  .412,  .42}38.40 & \cellcolor[rgb]{ .973,  .914,  .518}86.31 & \cellcolor[rgb]{ .945,  .906,  .518}77.90 & \cellcolor[rgb]{ .973,  .439,  .424}59.17 & \cellcolor[rgb]{ .976,  .51,  .435}48.39 & \cellcolor[rgb]{ .843,  .878,  .51}82.29 & \cellcolor[rgb]{ .859,  .882,  .51}81.32 & \cellcolor[rgb]{ .98,  .6,  .455}69.80 & \cellcolor[rgb]{ .984,  .647,  .463}62.63 & \cellcolor[rgb]{ .973,  .412,  .42}64.92 & \cellcolor[rgb]{ .973,  .412,  .42}61.06 & \cellcolor[rgb]{ .984,  .663,  .467}72.14 & \cellcolor[rgb]{ .984,  .655,  .467}65.39 & \cellcolor[rgb]{ .922,  .898,  .514}62.50 & \cellcolor[rgb]{ .863,  .882,  .51}58.85 & \cellcolor[rgb]{ .973,  .412,  .42}56.80 & \cellcolor[rgb]{ .973,  .412,  .42}56.49 & \cellcolor[rgb]{ .984,  .651,  .463}68.71 & \cellcolor[rgb]{ .984,  .651,  .463}68.55 \\
          & A-GEM & \cellcolor[rgb]{ .831,  .875,  .51}86.06 & \cellcolor[rgb]{ .749,  .851,  .506}78.44 & \cellcolor[rgb]{ .973,  .412,  .42}79.74 & \cellcolor[rgb]{ .973,  .412,  .42}71.78 & \cellcolor[rgb]{ .973,  .443,  .424}59.33 & \cellcolor[rgb]{ .973,  .467,  .427}45.94 & \cellcolor[rgb]{ .996,  .886,  .51}80.23 & \cellcolor[rgb]{ .996,  .886,  .51}79.27 & \cellcolor[rgb]{ .98,  .62,  .459}70.67 & \cellcolor[rgb]{ .984,  .635,  .463}61.77 & \cellcolor[rgb]{ .776,  .859,  .506}87.53 & \cellcolor[rgb]{ .8,  .867,  .51}86.61 & \cellcolor[rgb]{ .494,  .776,  .49}93.31 & \cellcolor[rgb]{ .498,  .776,  .49}92.95 & \cellcolor[rgb]{ .992,  .824,  .498}58.50 & \cellcolor[rgb]{ .988,  .729,  .478}46.86 & \cellcolor[rgb]{ .965,  .914,  .518}71.95 & \cellcolor[rgb]{ .933,  .902,  .514}71.89 & \cellcolor[rgb]{ .91,  .898,  .514}72.49 & \cellcolor[rgb]{ .906,  .894,  .514}72.34 \\
          & DER++ & \cellcolor[rgb]{ .996,  .898,  .51}84.27 & \cellcolor[rgb]{ .996,  .894,  .51}75.08 & \cellcolor[rgb]{ .588,  .804,  .494}87.53 & \cellcolor[rgb]{ .694,  .835,  .502}80.09 & \cellcolor[rgb]{ .992,  .776,  .486}72.29 & \cellcolor[rgb]{ .992,  .812,  .494}66.28 & \cellcolor[rgb]{ .714,  .839,  .502}83.26 & \cellcolor[rgb]{ .737,  .847,  .506}82.29 & \cellcolor[rgb]{ .553,  .792,  .494}86.70 & \cellcolor[rgb]{ .569,  .8,  .494}85.46 & \cellcolor[rgb]{ .918,  .898,  .514}86.87 & \cellcolor[rgb]{ .929,  .902,  .514}85.99 & \cellcolor[rgb]{ .973,  .412,  .42}60.44 & \cellcolor[rgb]{ .973,  .412,  .42}49.67 & \cellcolor[rgb]{ .98,  .616,  .459}54.20 & \cellcolor[rgb]{ .976,  .549,  .443}39.22 & \cellcolor[rgb]{ .996,  .875,  .506}70.48 & \cellcolor[rgb]{ .996,  .867,  .506}69.98 & \cellcolor[rgb]{ .992,  .922,  .518}71.84 & \cellcolor[rgb]{ .988,  .918,  .518}71.69 \\
          & EWC   & \cellcolor[rgb]{ .737,  .847,  .506}86.37 & \cellcolor[rgb]{ .996,  .886,  .51}74.52 & \cellcolor[rgb]{ .878,  .89,  .514}86.60 & \cellcolor[rgb]{ .898,  .894,  .514}78.31 & \cellcolor[rgb]{ .537,  .788,  .494}82.38 & \cellcolor[rgb]{ .6,  .808,  .498}78.41 & \cellcolor[rgb]{ .894,  .894,  .514}81.88 & \cellcolor[rgb]{ .902,  .894,  .514}80.99 & \cellcolor[rgb]{ .984,  .671,  .467}72.77 & \cellcolor[rgb]{ .984,  .69,  .471}65.76 & \cellcolor[rgb]{ .996,  .902,  .514}85.78 & \cellcolor[rgb]{ .996,  .898,  .51}84.65 & \cellcolor[rgb]{ .992,  .839,  .502}80.26 & \cellcolor[rgb]{ .996,  .863,  .506}78.60 & \cellcolor[rgb]{ .388,  .745,  .482}\textbf{75.40} & \cellcolor[rgb]{ .396,  .749,  .486}71.98 & \cellcolor[rgb]{ .992,  .82,  .498}68.85 & \cellcolor[rgb]{ .992,  .804,  .494}68.18 & \cellcolor[rgb]{ .996,  .859,  .506}71.10 & \cellcolor[rgb]{ .996,  .859,  .506}70.90 \\
          & OWM   & \cellcolor[rgb]{ .541,  .788,  .494}87.02 & \cellcolor[rgb]{ .592,  .804,  .494}79.31 & \cellcolor[rgb]{ .996,  .91,  .514}86.11 & \cellcolor[rgb]{ .996,  .851,  .502}76.65 & \cellcolor[rgb]{ .973,  .412,  .42}58.07 & \cellcolor[rgb]{ .973,  .412,  .42}42.63 & \cellcolor[rgb]{ .996,  .886,  .51}80.25 & \cellcolor[rgb]{ .996,  .89,  .51}79.45 & \cellcolor[rgb]{ .604,  .808,  .498}86.30 & \cellcolor[rgb]{ .58,  .804,  .494}85.36 & \cellcolor[rgb]{ .992,  .82,  .498}82.31 & \cellcolor[rgb]{ .992,  .831,  .498}81.44 & \cellcolor[rgb]{ .969,  .914,  .518}84.54 & \cellcolor[rgb]{ .98,  .918,  .518}82.73 & \cellcolor[rgb]{ .973,  .412,  .42}50.00 & \cellcolor[rgb]{ .973,  .412,  .42}33.33 & \cellcolor[rgb]{ .902,  .894,  .514}72.23 & \cellcolor[rgb]{ .918,  .898,  .514}71.96 & \cellcolor[rgb]{ .973,  .412,  .42}65.97 & \cellcolor[rgb]{ .973,  .412,  .42}65.81 \\
          & UCL   & \cellcolor[rgb]{ .996,  .89,  .51}83.89 & \cellcolor[rgb]{ .996,  .89,  .51}74.82 & \cellcolor[rgb]{ .996,  .855,  .502}85.38 & \cellcolor[rgb]{ .996,  .875,  .506}76.90 & \cellcolor[rgb]{ .773,  .859,  .506}80.12 & \cellcolor[rgb]{ .894,  .894,  .514}74.13 & \cellcolor[rgb]{ .569,  .796,  .494}84.39 & \cellcolor[rgb]{ .569,  .8,  .494}83.59 & \cellcolor[rgb]{ .988,  .718,  .478}74.76 & \cellcolor[rgb]{ .988,  .745,  .482}69.48 & \cellcolor[rgb]{ .455,  .765,  .486}89.00 & \cellcolor[rgb]{ .447,  .765,  .486}88.27 & \cellcolor[rgb]{ .42,  .757,  .486}94.65 & \cellcolor[rgb]{ .416,  .753,  .486}94.63 & \cellcolor[rgb]{ .522,  .784,  .49}72.20 & \cellcolor[rgb]{ .569,  .8,  .494}67.15 & \cellcolor[rgb]{ .996,  .914,  .514}71.61 & \cellcolor[rgb]{ .996,  .914,  .514}71.46 & \cellcolor[rgb]{ .89,  .89,  .514}72.66 & \cellcolor[rgb]{ .886,  .89,  .514}72.50 \\
          & HAT   & \cellcolor[rgb]{ .624,  .816,  .498}86.74 & \cellcolor[rgb]{ .796,  .863,  .506}78.16 & \cellcolor[rgb]{ .992,  .804,  .494}84.73 & \cellcolor[rgb]{ .992,  .835,  .498}76.49 & \cellcolor[rgb]{ .839,  .878,  .51}79.48 & \cellcolor[rgb]{ .988,  .918,  .518}72.78 & \cellcolor[rgb]{ .996,  .878,  .506}79.98 & \cellcolor[rgb]{ .996,  .878,  .51}79.10 & \cellcolor[rgb]{ .478,  .773,  .49}87.29 & \cellcolor[rgb]{ .486,  .773,  .49}86.14 & \cellcolor[rgb]{ .659,  .824,  .498}88.07 & \cellcolor[rgb]{ .659,  .824,  .498}87.26 & \cellcolor[rgb]{ .482,  .773,  .49}93.51 & \cellcolor[rgb]{ .498,  .776,  .49}92.93 & \cellcolor[rgb]{ .98,  .596,  .455}53.80 & \cellcolor[rgb]{ .98,  .627,  .459}42.56 & \cellcolor[rgb]{ .996,  .851,  .502}69.78 & \cellcolor[rgb]{ .996,  .855,  .502}69.70 & \cellcolor[rgb]{ .996,  .859,  .502}71.06 & \cellcolor[rgb]{ .996,  .863,  .506}70.93 \\
          & CAT   & \cellcolor[rgb]{ .996,  .886,  .51}83.68 & \cellcolor[rgb]{ .992,  .808,  .494}68.64 & \cellcolor[rgb]{ .655,  .824,  .498}87.32 & \cellcolor[rgb]{ .545,  .792,  .494}81.42 & \cellcolor[rgb]{ .984,  .651,  .463}67.41 & \cellcolor[rgb]{ .984,  .643,  .463}56.22 & \cellcolor[rgb]{ .976,  .525,  .439}71.03 & \cellcolor[rgb]{ .976,  .529,  .439}69.40 & \cellcolor[rgb]{ .471,  .769,  .49}87.34 & \cellcolor[rgb]{ .439,  .761,  .486}86.51 & \cellcolor[rgb]{ .996,  .851,  .502}83.56 & \cellcolor[rgb]{ .996,  .859,  .506}82.74 & \cellcolor[rgb]{ .392,  .749,  .486}95.17 & \cellcolor[rgb]{ .392,  .749,  .486}95.16 & \cellcolor[rgb]{ .973,  .467,  .427}51.15 & \cellcolor[rgb]{ .984,  .663,  .467}44.11 & \cellcolor[rgb]{ .863,  .882,  .51}72.39 & \cellcolor[rgb]{ .89,  .89,  .514}72.08 & \cellcolor[rgb]{ .82,  .871,  .51}73.21 & \cellcolor[rgb]{ .812,  .867,  .51}73.10 \\
          & KAN   & \cellcolor[rgb]{ .996,  .918,  .514}85.49 & \cellcolor[rgb]{ .937,  .906,  .518}77.38 & \cellcolor[rgb]{ .984,  .682,  .471}83.20 & \cellcolor[rgb]{ .98,  .569,  .447}73.52 & \cellcolor[rgb]{ .996,  .902,  .514}77.27 & \cellcolor[rgb]{ .996,  .918,  .514}72.34 & \cellcolor[rgb]{ .973,  .412,  .42}68.02 & \cellcolor[rgb]{ .973,  .412,  .42}66.13 & \cellcolor[rgb]{ .996,  .898,  .514}82.32 & \cellcolor[rgb]{ .996,  .91,  .514}81.23 & \cellcolor[rgb]{ .996,  .91,  .514}86.10 & \cellcolor[rgb]{ .996,  .914,  .514}85.30 & \cellcolor[rgb]{ .984,  .682,  .471}73.07 & \cellcolor[rgb]{ .988,  .725,  .478}69.97 & \cellcolor[rgb]{ .82,  .871,  .51}64.95 & \cellcolor[rgb]{ .863,  .882,  .51}58.84 & \cellcolor[rgb]{ .78,  .859,  .506}72.77 & \cellcolor[rgb]{ .749,  .851,  .506}72.72 & \cellcolor[rgb]{ .722,  .843,  .502}73.97 & \cellcolor[rgb]{ .714,  .839,  .502}73.86 \\
          & SRK   & \cellcolor[rgb]{ .996,  .906,  .514}84.76 & \cellcolor[rgb]{ .733,  .847,  .506}78.52 & \cellcolor[rgb]{ .988,  .737,  .482}83.91 & \cellcolor[rgb]{ .984,  .647,  .463}74.38 & \cellcolor[rgb]{ .933,  .902,  .514}78.58 & \cellcolor[rgb]{ .765,  .855,  .506}76.03 & \cellcolor[rgb]{ .988,  .706,  .475}75.58 & \cellcolor[rgb]{ .988,  .718,  .478}74.64 & \cellcolor[rgb]{ .898,  .894,  .514}83.99 & \cellcolor[rgb]{ .914,  .898,  .514}82.66 & \cellcolor[rgb]{ .643,  .82,  .498}88.14 & \cellcolor[rgb]{ .62,  .812,  .498}87.45 & \cellcolor[rgb]{ .992,  .827,  .498}79.64 & \cellcolor[rgb]{ .996,  .851,  .502}77.89 & \cellcolor[rgb]{ .992,  .788,  .49}57.80 & \cellcolor[rgb]{ .992,  .831,  .498}51.26 & \cellcolor[rgb]{ .996,  .855,  .502}69.84 & \cellcolor[rgb]{ .996,  .859,  .506}69.79 & \cellcolor[rgb]{ .992,  .824,  .498}70.69 & \cellcolor[rgb]{ .992,  .827,  .498}70.54 \\
          & HOP (ours) & \cellcolor[rgb]{ .388,  .745,  .482}\textbf{87.51} & \cellcolor[rgb]{ .388,  .745,  .482}\textbf{80.45} & \cellcolor[rgb]{ .388,  .745,  .482}\textbf{88.16} & \cellcolor[rgb]{ .388,  .745,  .482}\textbf{82.76} & \cellcolor[rgb]{ .388,  .745,  .482}\textbf{83.79} & \cellcolor[rgb]{ .388,  .745,  .482}\textbf{81.45} & \cellcolor[rgb]{ .388,  .745,  .482}\textbf{85.74} & \cellcolor[rgb]{ .388,  .745,  .482}\textbf{84.98} & \cellcolor[rgb]{ .388,  .745,  .482}\textbf{87.98} & \cellcolor[rgb]{ .388,  .745,  .482}\textbf{86.91} & \cellcolor[rgb]{ .388,  .745,  .482}\textbf{89.30} & \cellcolor[rgb]{ .388,  .745,  .482}\textbf{88.54} & \cellcolor[rgb]{ .388,  .745,  .482}\textbf{95.23} & \cellcolor[rgb]{ .388,  .745,  .482}\textbf{95.20} & \cellcolor[rgb]{ .435,  .761,  .486}74.32 & \cellcolor[rgb]{ .388,  .745,  .482}\textbf{72.10} & \cellcolor[rgb]{ .388,  .745,  .482}\textbf{74.49} & \cellcolor[rgb]{ .388,  .745,  .482}\textbf{74.34} & \cellcolor[rgb]{ .388,  .745,  .482}\textbf{76.59} & \cellcolor[rgb]{ .388,  .745,  .482}\textbf{76.45} \\
   \midrule
   
    \multirow{13}[0]{*}{\begin{sideways}\textbf{Adapter-BERT}\end{sideways}} & SDL   & 85.96 & 78.07 & 85.96 & 78.07 & 76.31 & 71.04 & 76.31 & 71.04 & 88.30 & 87.31 & 88.30 & 87.31 & 96.20 & 96.19 & 96.20 & 96.19 & 72.54 & 71.26 & 72.54 & 71.26 \\\cdashline{2-22}
          & FT    & \cellcolor[rgb]{ .98,  .592,  .451}54.03 & \cellcolor[rgb]{ .984,  .682,  .471}44.81 & \cellcolor[rgb]{ .996,  .875,  .506}86.67 & \cellcolor[rgb]{ .996,  .906,  .514}78.04 & \cellcolor[rgb]{ .976,  .553,  .447}55.19 & \cellcolor[rgb]{ .973,  .451,  .427}35.28 & \cellcolor[rgb]{ .996,  .855,  .502}82.22 & \cellcolor[rgb]{ .996,  .859,  .506}81.32 & \cellcolor[rgb]{ .98,  .616,  .459}64.94 & \cellcolor[rgb]{ .98,  .624,  .459}63.40 & \cellcolor[rgb]{ .82,  .871,  .51}88.77 & \cellcolor[rgb]{ .824,  .871,  .51}88.21 & \cellcolor[rgb]{ .996,  .863,  .506}68.29 & \cellcolor[rgb]{ .996,  .847,  .502}61.70 & \cellcolor[rgb]{ .996,  .918,  .514}63.10 & \cellcolor[rgb]{ .996,  .867,  .506}57.87 & \cellcolor[rgb]{ .973,  .412,  .42}69.31 & \cellcolor[rgb]{ .973,  .412,  .42}67.80 & \cellcolor[rgb]{ .678,  .831,  .502}81.63 & \cellcolor[rgb]{ .631,  .816,  .498}81.59 \\
          & L2    & \cellcolor[rgb]{ .992,  .812,  .494}63.97 & \cellcolor[rgb]{ .992,  .804,  .494}52.43 & \cellcolor[rgb]{ .976,  .541,  .443}75.64 & \cellcolor[rgb]{ .984,  .671,  .467}59.78 & \cellcolor[rgb]{ .996,  .886,  .51}70.87 & \cellcolor[rgb]{ .996,  .902,  .514}69.11 & \cellcolor[rgb]{ .996,  .882,  .51}83.78 & \cellcolor[rgb]{ .996,  .882,  .51}82.86 & \cellcolor[rgb]{ .992,  .831,  .498}73.03 & \cellcolor[rgb]{ .996,  .851,  .502}71.50 & \cellcolor[rgb]{ .725,  .843,  .502}89.02 & \cellcolor[rgb]{ .722,  .843,  .502}88.49 & \cellcolor[rgb]{ .996,  .898,  .51}69.56 & \cellcolor[rgb]{ .996,  .914,  .514}65.50 & \cellcolor[rgb]{ .957,  .91,  .518}63.85 & \cellcolor[rgb]{ .941,  .906,  .518}61.29 & \cellcolor[rgb]{ .867,  .882,  .51}77.16 & \cellcolor[rgb]{ .863,  .882,  .51}76.85 & \cellcolor[rgb]{ .98,  .624,  .459}78.81 & \cellcolor[rgb]{ .984,  .686,  .471}78.91 \\
          & A-GEM & \cellcolor[rgb]{ .973,  .412,  .42}45.88 & \cellcolor[rgb]{ .973,  .412,  .42}28.21 & \cellcolor[rgb]{ .996,  .878,  .506}86.78 & \cellcolor[rgb]{ .996,  .906,  .514}77.84 & \cellcolor[rgb]{ .984,  .643,  .463}59.35 & \cellcolor[rgb]{ .988,  .702,  .475}54.20 & \cellcolor[rgb]{ .91,  .898,  .514}86.25 & \cellcolor[rgb]{ .882,  .89,  .514}85.49 & \cellcolor[rgb]{ .992,  .784,  .49}71.22 & \cellcolor[rgb]{ .992,  .804,  .494}69.94 & \cellcolor[rgb]{ .98,  .565,  .447}86.54 & \cellcolor[rgb]{ .973,  .412,  .42}56.12 & \cellcolor[rgb]{ .984,  .643,  .463}60.29 & \cellcolor[rgb]{ .984,  .655,  .467}50.40 & \cellcolor[rgb]{ .996,  .882,  .51}62.24 & \cellcolor[rgb]{ .996,  .918,  .514}60.30 & \cellcolor[rgb]{ .69,  .835,  .502}78.84 & \cellcolor[rgb]{ .745,  .851,  .506}78.04 & \cellcolor[rgb]{ .973,  .412,  .42}77.45 & \cellcolor[rgb]{ .973,  .475,  .431}77.36 \\
          & DER++ & \cellcolor[rgb]{ .973,  .447,  .424}47.63 & \cellcolor[rgb]{ .976,  .529,  .439}35.54 & \cellcolor[rgb]{ .871,  .886,  .514}88.59 & \cellcolor[rgb]{ .918,  .898,  .514}79.85 & \cellcolor[rgb]{ .988,  .722,  .478}63.11 & \cellcolor[rgb]{ .992,  .808,  .494}61.96 & \cellcolor[rgb]{ .996,  .886,  .51}84.00 & \cellcolor[rgb]{ .996,  .89,  .51}83.16 & \cellcolor[rgb]{ .973,  .478,  .431}59.67 & \cellcolor[rgb]{ .973,  .467,  .427}57.82 & \cellcolor[rgb]{ .984,  .682,  .471}87.12 & \cellcolor[rgb]{ .996,  .898,  .51}86.43 & \cellcolor[rgb]{ .98,  .608,  .455}58.95 & \cellcolor[rgb]{ .984,  .639,  .463}49.58 & \cellcolor[rgb]{ .992,  .804,  .494}60.18 & \cellcolor[rgb]{ .996,  .867,  .506}57.89 & \cellcolor[rgb]{ .992,  .784,  .49}74.12 & \cellcolor[rgb]{ .992,  .812,  .494}73.78 & \cellcolor[rgb]{ .992,  .784,  .49}79.84 & \cellcolor[rgb]{ .992,  .776,  .486}79.56 \\
          & EWC   & \cellcolor[rgb]{ .984,  .643,  .463}56.30 & \cellcolor[rgb]{ .988,  .757,  .486}49.58 & \cellcolor[rgb]{ .996,  .918,  .514}88.05 & \cellcolor[rgb]{ .996,  .918,  .514}78.75 & \cellcolor[rgb]{ .98,  .62,  .459}58.34 & \cellcolor[rgb]{ .976,  .553,  .447}42.85 & \cellcolor[rgb]{ .569,  .8,  .494}87.29 & \cellcolor[rgb]{ .596,  .808,  .498}86.34 & \cellcolor[rgb]{ .98,  .557,  .447}62.69 & \cellcolor[rgb]{ .98,  .569,  .447}61.51 & \cellcolor[rgb]{ .886,  .89,  .514}88.58 & \cellcolor[rgb]{ .922,  .902,  .514}87.94 & \cellcolor[rgb]{ .984,  .686,  .471}61.86 & \cellcolor[rgb]{ .988,  .714,  .475}53.94 & \cellcolor[rgb]{ .973,  .412,  .42}49.95 & \cellcolor[rgb]{ .973,  .412,  .42}36.86 & \cellcolor[rgb]{ .996,  .91,  .514}75.72 & \cellcolor[rgb]{ .996,  .918,  .514}75.38 & \cellcolor[rgb]{ .996,  .91,  .514}80.63 & \cellcolor[rgb]{ .996,  .918,  .514}80.59 \\
          & OWM   & \cellcolor[rgb]{ .878,  .886,  .514}72.99 & \cellcolor[rgb]{ .835,  .875,  .51}66.51 & \cellcolor[rgb]{ .996,  .906,  .514}87.66 & \cellcolor[rgb]{ .996,  .918,  .514}78.82 & \cellcolor[rgb]{ .929,  .902,  .514}73.97 & \cellcolor[rgb]{ .937,  .906,  .518}71.96 & \cellcolor[rgb]{ .988,  .769,  .486}77.32 & \cellcolor[rgb]{ .992,  .784,  .49}76.68 & \cellcolor[rgb]{ .596,  .808,  .498}85.46 & \cellcolor[rgb]{ .584,  .804,  .494}84.57 & \cellcolor[rgb]{ .973,  .412,  .42}85.78 & \cellcolor[rgb]{ .996,  .882,  .51}85.45 & \cellcolor[rgb]{ .984,  .918,  .518}71.10 & \cellcolor[rgb]{ .996,  .922,  .518}66.25 & \cellcolor[rgb]{ .992,  .804,  .494}60.23 & \cellcolor[rgb]{ .996,  .867,  .506}57.94 & \cellcolor[rgb]{ .976,  .498,  .435}70.45 & \cellcolor[rgb]{ .973,  .439,  .424}68.24 & \cellcolor[rgb]{ .973,  .439,  .424}77.65 & \cellcolor[rgb]{ .973,  .412,  .42}76.89 \\
          & UCL   & \cellcolor[rgb]{ .992,  .824,  .498}64.46 & \cellcolor[rgb]{ .976,  .549,  .443}36.64 & \cellcolor[rgb]{ .973,  .412,  .42}71.23 & \cellcolor[rgb]{ .973,  .412,  .42}39.61 & \cellcolor[rgb]{ .973,  .412,  .42}48.36 & \cellcolor[rgb]{ .973,  .412,  .42}32.07 & \cellcolor[rgb]{ .973,  .412,  .42}56.53 & \cellcolor[rgb]{ .973,  .412,  .42}52.96 & \cellcolor[rgb]{ .973,  .412,  .42}57.06 & \cellcolor[rgb]{ .973,  .412,  .42}55.86 & \cellcolor[rgb]{ .996,  .902,  .514}88.18 & \cellcolor[rgb]{ .996,  .918,  .514}87.54 & \cellcolor[rgb]{ .973,  .412,  .42}51.75 & \cellcolor[rgb]{ .973,  .412,  .42}36.06 & \cellcolor[rgb]{ .973,  .42,  .42}50.25 & \cellcolor[rgb]{ .973,  .412,  .42}36.78 & \cellcolor[rgb]{ .929,  .902,  .514}76.54 & \cellcolor[rgb]{ .902,  .894,  .514}76.43 & \cellcolor[rgb]{ .996,  .867,  .506}80.36 & \cellcolor[rgb]{ .996,  .89,  .51}80.40 \\
          & HAT   & \cellcolor[rgb]{ .498,  .776,  .49}86.14 & \cellcolor[rgb]{ .545,  .792,  .494}78.52 & \cellcolor[rgb]{ .976,  .918,  .518}88.23 & \cellcolor[rgb]{ .984,  .918,  .518}79.19 & \cellcolor[rgb]{ .612,  .812,  .498}80.83 & \cellcolor[rgb]{ .647,  .824,  .498}78.41 & \cellcolor[rgb]{ .918,  .898,  .514}86.22 & \cellcolor[rgb]{ .898,  .894,  .514}85.44 & \cellcolor[rgb]{ .482,  .773,  .49}88.00 & \cellcolor[rgb]{ .478,  .773,  .49}87.26 & \cellcolor[rgb]{ .98,  .561,  .447}86.52 & \cellcolor[rgb]{ .996,  .89,  .51}85.78 & \cellcolor[rgb]{ .392,  .749,  .486}95.22 & \cellcolor[rgb]{ .392,  .749,  .486}95.21 & \cellcolor[rgb]{ .996,  .922,  .518}63.30 & \cellcolor[rgb]{ .996,  .922,  .518}60.44 & \cellcolor[rgb]{ .98,  .58,  .451}71.51 & \cellcolor[rgb]{ .973,  .478,  .431}68.83 & \cellcolor[rgb]{ .894,  .89,  .514}81.02 & \cellcolor[rgb]{ .886,  .89,  .514}80.92 \\
          & B-CL  & \cellcolor[rgb]{ .435,  .761,  .486}88.29 & \cellcolor[rgb]{ .478,  .773,  .49}81.40 & \cellcolor[rgb]{ .506,  .78,  .49}89.83 & \cellcolor[rgb]{ .478,  .773,  .49}\textbf{84.22} & \cellcolor[rgb]{ .451,  .765,  .486}84.34 & \cellcolor[rgb]{ .439,  .761,  .486}83.12 & \cellcolor[rgb]{ .996,  .918,  .514}85.92 & \cellcolor[rgb]{ .996,  .918,  .514}85.11 & \cellcolor[rgb]{ .851,  .878,  .51}79.76 & \cellcolor[rgb]{ .902,  .894,  .514}76.51 & \cellcolor[rgb]{ .996,  .89,  .51}88.12 & \cellcolor[rgb]{ .996,  .918,  .514}87.48 & \cellcolor[rgb]{ .396,  .749,  .486}95.07 & \cellcolor[rgb]{ .396,  .749,  .486}95.04 & \cellcolor[rgb]{ .914,  .898,  .514}64.50 & \cellcolor[rgb]{ .902,  .894,  .514}61.87 & \cellcolor[rgb]{ .984,  .69,  .471}72.92 & \cellcolor[rgb]{ .988,  .737,  .482}72.71 & \cellcolor[rgb]{ .82,  .871,  .51}81.23 & \cellcolor[rgb]{ .847,  .878,  .51}81.02 \\
          & CTR   & \cellcolor[rgb]{ .4,  .749,  .486}89.47 & \cellcolor[rgb]{ .424,  .757,  .486}83.62 & \cellcolor[rgb]{ .71,  .839,  .502}89.13 & \cellcolor[rgb]{ .549,  .792,  .494}83.52 & \cellcolor[rgb]{ .467,  .769,  .49}83.96 & \cellcolor[rgb]{ .443,  .761,  .486}83.00 & \cellcolor[rgb]{ .988,  .918,  .518}86.01 & \cellcolor[rgb]{ .992,  .922,  .518}85.16 & \cellcolor[rgb]{ .424,  .757,  .486}89.31 & \cellcolor[rgb]{ .416,  .757,  .486}88.75 & \cellcolor[rgb]{ .969,  .914,  .518}88.36 & \cellcolor[rgb]{ .937,  .906,  .518}87.89 & \cellcolor[rgb]{ .392,  .749,  .486}95.25 & \cellcolor[rgb]{ .392,  .749,  .486}95.23 & \cellcolor[rgb]{ .831,  .875,  .51}65.76 & \cellcolor[rgb]{ .827,  .875,  .51}63.04 & \cellcolor[rgb]{ .988,  .922,  .518}75.95 & \cellcolor[rgb]{ .996,  .922,  .518}75.46 & \cellcolor[rgb]{ .976,  .918,  .518}80.78 & \cellcolor[rgb]{ .992,  .922,  .518}80.64 \\
          & HOP (ours) & \cellcolor[rgb]{ .388,  .745,  .482}\textbf{89.84} & \cellcolor[rgb]{ .388,  .745,  .482}\textbf{85.06} & \cellcolor[rgb]{ .494,  .776,  .49}\textbf{89.87} & \cellcolor[rgb]{ .49,  .776,  .49}84.12 & \cellcolor[rgb]{ .388,  .745,  .482}\textbf{85.63} & \cellcolor[rgb]{ .388,  .745,  .482}\textbf{84.18} & \cellcolor[rgb]{ .388,  .745,  .482}\textbf{87.84} & \cellcolor[rgb]{ .388,  .745,  .482}\textbf{86.96} & \cellcolor[rgb]{ .388,  .745,  .482}\textbf{90.08} & \cellcolor[rgb]{ .388,  .745,  .482}\textbf{89.44} & \cellcolor[rgb]{ .388,  .745,  .482}\textbf{89.93} & \cellcolor[rgb]{ .388,  .745,  .482}\textbf{89.41} & \cellcolor[rgb]{ .388,  .745,  .482}\textbf{95.30} & \cellcolor[rgb]{ .388,  .745,  .482}\textbf{95.29} & \cellcolor[rgb]{ .388,  .745,  .482}\textbf{72.30} & \cellcolor[rgb]{ .388,  .745,  .482}\textbf{69.67} & \cellcolor[rgb]{ .388,  .745,  .482}\textbf{81.75} & \cellcolor[rgb]{ .388,  .745,  .482}\textbf{81.66} & \cellcolor[rgb]{ .388,  .745,  .482}\textbf{82.46} & \cellcolor[rgb]{ .388,  .745,  .482}\textbf{82.22} \\
          \bottomrule
    \end{tabular}%
    }
  \label{tab:main_results}%
  \vspace{-0.4cm}
\end{table*}%

\noindent\textbf{Metrics.} We compute both mean accuracy (mAcc, $\uparrow$) and macro-F1 (MF1, $\uparrow$), to reduce biases in accuracy originating from imbalanced classes.
To fully characterize the different approaches, we report a wide range of forgetting and transfer metrics computed from the accuracy matrix \cite{mai2022online}, as illustrated in Figure~\ref{fig:CL_metrics}.
Namely, we report: backward transfer ($BwT$, $\uparrow$), which tracks the influence that learning a new problem has on the preceding problems’ performance, to measure stability; 
forward transfer ($FwT$, $\uparrow$) measures the positive influence of learning a problem on future problems’ performance;
forgetting ($Forg$, $\downarrow$) averages the difference of class-wise accuracy achieved at the last step and the best class-wise accuracy achieved previously;
plasticity ($Pla$, $\uparrow$) averages the accuracy achieved on each problem evaluated right after learning that problem.
Additionally, we report the number of overall (\#OP, $\downarrow$) and of trainable parameters (\#TP, $\downarrow$), and the computation time ($\downarrow$, in minutes). %

\section{Experimental Results}
\label{sec:results}

\begin{table}[t]
\setlength{\tabcolsep}{1.2pt}
  \centering
  \caption{Aggregate results from Tab.~\ref{tab:main_results}. First (second) vertical group of 2 columns: results averaged over the 5 benchmarks for TIL (DIL). Last vertical group: averaged over both benchmarks and CL setups. }
  \resizebox{0.75\linewidth}{!}{%
    \begin{tabular}{clcccccc}
    \toprule
    \multirow{3}[0]{*}{} & \multicolumn{1}{c}{\multirow{3}[0]{*}{\textbf{CL Method}}} & \multicolumn{6}{c}{\textbf{Avg Benchmarks}} \\ \cmidrule(lr){3-8}
          &       & \multicolumn{2}{c}{\textbf{TIL}} & \multicolumn{2}{c}{\textbf{DIL}} & \multicolumn{2}{c}{\textbf{Avg Setups}} \\\cmidrule(lr){3-4}\cmidrule(lr){5-6}\cmidrule(lr){7-8}
          &       & $\mathbf{mAcc}$  & $\mathbf{MF1}$ & $\mathbf{mAcc}$  & $\mathbf{MF1}$ & $\mathbf{mAcc}$  & $\mathbf{MF1}$ \\
          \midrule
    \multirow{3}[0]{*}{\begin{sideways}\textbf{BF+Lin}\end{sideways}} & SDL   & 66.32 & 57.84 & 58.40 & 53.00 & 62.36 & 55.42 \\\cdashline{2-8}
          & FT    & 60.19 & 50.13 & 61.31 & 56.45 & 60.75 & 53.29 \\
          & HOP (ours) & \textbf{81.39} & \textbf{78.89} & \textbf{76.56} & \textbf{74.47} & \textbf{78.98} & \textbf{76.68} \\\midrule
    \multirow{13}[0]{*}{\begin{sideways}\textbf{Bert (frozen) + CNN}\end{sideways}} & SDL   & 80.41 & 74.05 & 80.41 & 74.05 & 80.41 & 74.05 \\\cdashline{2-8}
          & FT    & \cellcolor[rgb]{ .996,  .886,  .51}77.15 & \cellcolor[rgb]{ .992,  .843,  .502}71.03 & \cellcolor[rgb]{ .729,  .843,  .502}78.84 & \cellcolor[rgb]{ .737,  .847,  .506}75.67 & \cellcolor[rgb]{ .937,  .906,  .518}78.00 & \cellcolor[rgb]{ .996,  .922,  .518}73.35 \\
          & L2    & \cellcolor[rgb]{ .973,  .412,  .42}62.79 & \cellcolor[rgb]{ .973,  .412,  .42}54.26 & \cellcolor[rgb]{ .973,  .412,  .42}72.95 & \cellcolor[rgb]{ .984,  .675,  .467}69.54 & \cellcolor[rgb]{ .973,  .412,  .42}67.87 & \cellcolor[rgb]{ .973,  .412,  .42}61.90 \\
          & A-GEM & \cellcolor[rgb]{ .996,  .859,  .502}76.26 & \cellcolor[rgb]{ .992,  .824,  .498}70.20 & \cellcolor[rgb]{ .996,  .922,  .518}75.70 & \cellcolor[rgb]{ .996,  .894,  .51}71.37 & \cellcolor[rgb]{ .996,  .851,  .502}75.98 & \cellcolor[rgb]{ .992,  .808,  .494}70.79 \\
          & DER++ & \cellcolor[rgb]{ .992,  .808,  .494}74.84 & \cellcolor[rgb]{ .992,  .8,  .494}69.29 & \cellcolor[rgb]{ .906,  .894,  .514}76.74 & \cellcolor[rgb]{ .984,  .918,  .518}71.86 & \cellcolor[rgb]{ .992,  .839,  .502}75.79 & \cellcolor[rgb]{ .992,  .8,  .494}70.58 \\
          & EWC   & \cellcolor[rgb]{ .996,  .918,  .514}78.13 & \cellcolor[rgb]{ .996,  .898,  .51}73.09 & \cellcolor[rgb]{ .616,  .812,  .498}80.15 & \cellcolor[rgb]{ .624,  .816,  .498}77.37 & \cellcolor[rgb]{ .839,  .878,  .51}79.14 & \cellcolor[rgb]{ .867,  .886,  .514}75.23 \\
          & OWM   & \cellcolor[rgb]{ .996,  .902,  .514}77.63 & \cellcolor[rgb]{ .996,  .878,  .51}72.40 & \cellcolor[rgb]{ .973,  .412,  .42}72.93 & \cellcolor[rgb]{ .973,  .412,  .42}67.34 & \cellcolor[rgb]{ .992,  .812,  .494}75.28 & \cellcolor[rgb]{ .988,  .769,  .486}69.87 \\
          & UCL   & \cellcolor[rgb]{ .773,  .859,  .506}81.01 & \cellcolor[rgb]{ .816,  .871,  .51}76.90 & \cellcolor[rgb]{ .569,  .8,  .494}80.73 & \cellcolor[rgb]{ .604,  .808,  .498}77.68 & \cellcolor[rgb]{ .69,  .831,  .502}80.87 & \cellcolor[rgb]{ .729,  .843,  .502}77.29 \\
          & HAT   & \cellcolor[rgb]{ .584,  .804,  .494}83.36 & \cellcolor[rgb]{ .624,  .816,  .498}79.94 & \cellcolor[rgb]{ .996,  .902,  .514}75.53 & \cellcolor[rgb]{ .996,  .882,  .51}71.27 & \cellcolor[rgb]{ .812,  .871,  .51}79.44 & \cellcolor[rgb]{ .843,  .878,  .51}75.61 \\
          & CAT   & \cellcolor[rgb]{ .757,  .855,  .506}81.20 & \cellcolor[rgb]{ .89,  .89,  .514}75.72 & \cellcolor[rgb]{ .973,  .471,  .431}73.25 & \cellcolor[rgb]{ .988,  .749,  .482}70.15 & \cellcolor[rgb]{ .996,  .918,  .514}77.23 & \cellcolor[rgb]{ .996,  .906,  .514}72.94 \\
          & KAN   & \cellcolor[rgb]{ 1,  .922,  .518}78.18 & \cellcolor[rgb]{ .949,  .91,  .518}74.73 & \cellcolor[rgb]{ .996,  .851,  .502}75.25 & \cellcolor[rgb]{ .996,  .914,  .514}71.53 & \cellcolor[rgb]{ .996,  .89,  .51}76.72 & \cellcolor[rgb]{ .996,  .914,  .514}73.13 \\
          & SRK   & \cellcolor[rgb]{ .906,  .894,  .514}79.36 & \cellcolor[rgb]{ .808,  .867,  .51}76.98 & \cellcolor[rgb]{ .992,  .847,  .502}75.22 & \cellcolor[rgb]{ .996,  .922,  .518}71.65 & \cellcolor[rgb]{ 1,  .922,  .518}77.29 & \cellcolor[rgb]{ .929,  .902,  .514}74.32 \\
          & HOP (ours) & \cellcolor[rgb]{ .388,  .745,  .482}\textbf{85.80} & \cellcolor[rgb]{ .388,  .745,  .482}\textbf{83.67} & \cellcolor[rgb]{ .388,  .745,  .482}\textbf{82.82} & \cellcolor[rgb]{ .388,  .745,  .482}\textbf{80.97} & \cellcolor[rgb]{ .388,  .745,  .482}\textbf{84.31} & \cellcolor[rgb]{ .388,  .745,  .482}\textbf{82.32} \\\midrule
    \multirow{13}[0]{*}{\begin{sideways}\textbf{Adapter-BERT}\end{sideways}} & SDL   & 83.86 & 80.77 & 83.86 & 80.77 & 83.86 & 80.77 \\\cdashline{2-8}
          & FT    & \cellcolor[rgb]{ .976,  .514,  .439}62.35 & \cellcolor[rgb]{ .98,  .576,  .451}54.60 & \cellcolor[rgb]{ .965,  .914,  .518}80.48 & \cellcolor[rgb]{ 1,  .922,  .518}77.41 & \cellcolor[rgb]{ .988,  .733,  .478}71.42 & \cellcolor[rgb]{ .988,  .753,  .482}66.00 \\
          & L2    & \cellcolor[rgb]{ .992,  .843,  .502}70.92 & \cellcolor[rgb]{ .996,  .867,  .506}67.08 & \cellcolor[rgb]{ .992,  .827,  .498}78.22 & \cellcolor[rgb]{ .992,  .831,  .498}74.27 & \cellcolor[rgb]{ .996,  .878,  .51}74.57 & \cellcolor[rgb]{ .996,  .882,  .51}70.67 \\
          & A-GEM & \cellcolor[rgb]{ .976,  .545,  .443}63.12 & \cellcolor[rgb]{ .98,  .612,  .455}56.16 & \cellcolor[rgb]{ .996,  .902,  .514}79.85 & \cellcolor[rgb]{ .988,  .749,  .482}71.42 & \cellcolor[rgb]{ .988,  .737,  .482}71.48 & \cellcolor[rgb]{ .984,  .694,  .471}63.79 \\
          & DER++ & \cellcolor[rgb]{ .973,  .451,  .427}60.70 & \cellcolor[rgb]{ .98,  .604,  .455}55.74 & \cellcolor[rgb]{ .996,  .906,  .514}79.95 & \cellcolor[rgb]{ .996,  .918,  .514}77.38 & \cellcolor[rgb]{ .984,  .682,  .471}70.32 & \cellcolor[rgb]{ .988,  .769,  .486}66.56 \\
          & EWC   & \cellcolor[rgb]{ .976,  .537,  .443}62.98 & \cellcolor[rgb]{ .98,  .624,  .459}56.65 & \cellcolor[rgb]{ .996,  .859,  .502}78.90 & \cellcolor[rgb]{ .992,  .827,  .498}74.10 & \cellcolor[rgb]{ .988,  .71,  .475}70.94 & \cellcolor[rgb]{ .988,  .737,  .482}65.37 \\
          & OWM   & \cellcolor[rgb]{ .925,  .902,  .514}74.79 & \cellcolor[rgb]{ .925,  .902,  .514}71.51 & \cellcolor[rgb]{ .992,  .804,  .494}77.73 & \cellcolor[rgb]{ .996,  .855,  .502}75.16 & \cellcolor[rgb]{ .957,  .91,  .518}76.26 & \cellcolor[rgb]{ .937,  .906,  .518}73.33 \\
          & UCL   & \cellcolor[rgb]{ .973,  .412,  .42}59.63 & \cellcolor[rgb]{ .973,  .412,  .42}47.41 & \cellcolor[rgb]{ .973,  .412,  .42}69.31 & \cellcolor[rgb]{ .973,  .412,  .42}59.46 & \cellcolor[rgb]{ .973,  .412,  .42}64.47 & \cellcolor[rgb]{ .973,  .412,  .42}53.44 \\
          & HAT   & \cellcolor[rgb]{ .553,  .796,  .494}84.34 & \cellcolor[rgb]{ .576,  .8,  .494}81.65 & \cellcolor[rgb]{ .882,  .89,  .514}81.06 & \cellcolor[rgb]{ .886,  .89,  .514}78.35 & \cellcolor[rgb]{ .6,  .808,  .498}82.70 & \cellcolor[rgb]{ .62,  .812,  .498}80.00 \\
          & B-CL  & \cellcolor[rgb]{ .565,  .796,  .494}84.08 & \cellcolor[rgb]{ .573,  .8,  .494}81.76 & \cellcolor[rgb]{ .757,  .851,  .506}81.92 & \cellcolor[rgb]{ .694,  .835,  .502}79.94 & \cellcolor[rgb]{ .584,  .804,  .494}83.00 & \cellcolor[rgb]{ .58,  .8,  .494}80.85 \\
          & CTR   & \cellcolor[rgb]{ .459,  .765,  .486}86.79 & \cellcolor[rgb]{ .455,  .765,  .486}85.21 & \cellcolor[rgb]{ .745,  .851,  .506}82.01 & \cellcolor[rgb]{ .682,  .831,  .502}80.05 & \cellcolor[rgb]{ .506,  .78,  .49}84.40 & \cellcolor[rgb]{ .494,  .776,  .49}82.63 \\
          & HOP (ours) & \cellcolor[rgb]{ .388,  .745,  .482}\textbf{88.52} & \cellcolor[rgb]{ .388,  .745,  .482}\textbf{87.13} & \cellcolor[rgb]{ .388,  .745,  .482}\textbf{84.48} & \cellcolor[rgb]{ .388,  .745,  .482}\textbf{82.48} & \cellcolor[rgb]{ .388,  .745,  .482}\textbf{86.50} & \cellcolor[rgb]{ .388,  .745,  .482}\textbf{84.80} \\
          \bottomrule
    \end{tabular}%
    }
  \label{tab:main_results_averaged}%
  \vspace{-0.5cm}
\end{table}%

\textbf{Main Results.}
We evaluate methods on 5 benchmarks (ASC, DSC small, DSC full, 20News, NLI) targeting 4 applications (ASC, DSC, TC, NLI) in 2 CL setups (DIL and TIL) using 3 network architectures based on BERT. 
Table~\ref{tab:main_results} shows that HOP clearly outperforms or achieves comparable results to baseline competitors in every scenario.

In the first block, we evaluate them on BF + Lin. Due to the low accuracy of this architecture, we compare our framework only against SDL and FT. HOP outperforms both by a large margin in every case.
In the second block, we use  BF + CNN \cite{kim2014convolutional}. Here, we report comparison against several approaches (we note that B-CL and CTR cannot be employed with a CNN head).
Finally, the best results are achieved on Adapter-BERT reported in the third block (we remark that CAT, KAN, SRK cannot work with adapters).
We observe that $mAcc$ and $MF1$ generally show consensus in identifying the best methods. Also, results are higher in DIL since a single head can transfer knowledge more easily.

SDL outperforms some approaches, due to increased model size for adaptation to  end problems. However, it builds a model for each problem independently using a separate network, therefore, it does not handle CF or KT.
On the other hand: FT,  regularization-based approaches (such as EWC, OWM, and L2) and replay-based approaches (such as A-GEM and DER++) perform generally better in BF+CNN than in Adapter-BERT, due to the fewer parameters used to update models and apply regularization on them.

KAN and HAT require problem identifier, and suffer from CF in TIL. We extended them to DIL by using the last model, which, however, shows low results in DIL. 
Similarly, also CAT (which extends HAT), SRK and UCL cannot achieve competitive results.
Approaches specifically designed for CL in NLP (\ie, B-CL and CTR) show clear improvements compared to the others.
B-CL and CTR have been mainly designed for TIL: they achieve competitive results in TIL setup, however they fail when employed in DIL.
HOP outperforms or it is comparable to the current state-of-the-art  competitors in every scenario, and it can deal both with large scale data (\eg, DSC full) and with limited data (\eg, DSC small) in both TIL and DIL. 
We confirm these findings by looking at the aggregate results reported in Table~\ref{tab:main_results_averaged}. The results show that HOP robustly outperforms all baseline competitors in both TIL and DIL (first and second vertical groups). Also, CIL-based methods are inadequate for TIL and DIL in NLP. 
Finally, the last vertical block provides a further comparison aggregated across all benchmarks and CL setups, and is helpful to grasp an overall sense of the results. 
In general, the best performing frameworks for BF+CNN are HOP, UCL and HAT; while for Adapter-BERT, they are HOP, CTR, B-CL, and HAT.

\begin{table*}[t]
\setlength{\tabcolsep}{2pt}
  \scriptsize
  \centering
  \caption{Collection of metrics on the DSC small dataset for every competing approach on both TIL and DIL setups.}
  \resizebox{0.9\linewidth}{!}
  {%
    \begin{tabular}{lcccccc:cccccc:ccc}
    \toprule
          & \multicolumn{6}{c}{\textbf{TIL}}              & \multicolumn{6}{:c:}{\textbf{DIL}}              &       &       &  \\\cmidrule(lr){2-7} \cmidrule(lr){8-13}
          & $\mathbf{mAcc} \uparrow$  & $\mathbf{MF1} \uparrow$ & $\mathbf{BwT} \uparrow$ & $\mathbf{FwT} \uparrow$ & $\mathbf{Forg} \downarrow$ & $\mathbf{Pla} \uparrow$ & $\mathbf{mAcc} \uparrow$  & $\mathbf{MF1} \uparrow$ & $\mathbf{BwT} \uparrow$ & $\mathbf{FwT} \uparrow$ & $\mathbf{Forg} \downarrow$ & $\mathbf{Pla} \uparrow$ & \textbf{\#OP} $\downarrow$ & \textbf{\#TP} $\downarrow$ & \textbf{Time} $\downarrow$ \\ \midrule
    {FT} & 55.19 & 35.28 & 59.59 & 45.50 & 28.79 & 74.71 & 82.22 & 81.32 & 83.17 & 79.24 & 5.56  & 82.99 & 183.3 & 73.8  & 1.2 \\
    {L2} & 70.87 & 69.11 & 71.48 & 57.46 & 13.60 & 82.86 & 83.78 & 82.86 & 83.48 & 81.22 & 1.44  & 82.99 & 183.3 & 73.8  & 1.4 \\
    {A-GEM}  & 59.35 & 54.20 & 56.77 & 46.14 & 26.05 & 84.24 & 86.25 & 85.49 & 84.92 & 83.03 & 0.90 & 84.80 & 183.3 & 73.8  & 8.7 \\
    {DER++}  & 63.11 & 61.96 & 50.92 & 40.26 & 22.76 & 83.31 & 84.00 & 83.16 & 85.06 & 80.97 & 3.46  & 84.15 & 183.3 & 73.8  & 1.2 \\
    {EWC}  & 58.34 & 42.85 & 54.15 & 43.10 & 25.65 & 79.13 & 87.29 & 86.34 & 86.05 & 82.25 & 0.83  & 83.19 & 183.3 & 73.8  & 1.3 \\
    {OWM}  & 73.97 & 71.96 & 72.19 & 58.32 & 12.40 & 83.39 & 77.32 & 76.68 & 73.78 & 54.12 & 6.76  & 79.87 & 184.4 & 74.8  & 1.4 \\
    {UCL}  & 48.36 & 32.07 & 51.19 & 49.21 & 7.48  & 52.12 & 56.53 & 52.96 & 51.07 & 43.16 & 7.08  & 57.71 & 183.4 & 73.9  & 1.3 \\
    {HAT}  & 80.83 & 78.41 & 75.97 & 58.15 & 6.50  & 83.12 & 86.22 & 85.44 & 85.25 & 80.57 & 1.57  & 84.26 & 184.0 & 74.5  & 1.6 \\\hdashline
    {B-CL}  & 84.34 & 83.12 & 84.76 & 48.40 & 0.67  & 81.98 & 85.92 & 85.11 & 85.18 & 78.73 & 1.34  & 85.92 & 220.2 & 110.7 & 4.8 \\
    {CTR} & 83.96 & 83.00 & 84.01 & 47.95 & 0.28  & 82.86 & 86.01 & 85.16 & 85.26 & 78.81 & 1.56  & 85.89 & 186.6 & 77.1  & 30.8 \\\hdashline
    {HOP (ours)} & 85.63 & 84.18 & 84.39 & 45.32 & 0.57  & 82.76 & \textbf{87.84} & \textbf{86.96} & 85.11 & 82.33 & 0.39  & 84.41 & 189.2 & 79.7  & 1.3 \\\hdashline
    {B-CL + HOP (ours)} & 86.93 & 86.25 & 86.84 & 51.47 & 0.24  & 86.53 & 87.08 & 86.30 & 86.47 & 84.49 & 0.88  & 87.22 & 244.8 & 134.3 & 4.9 \\
    {CTR + HOP (ours)} & \textbf{87.08} & \textbf{86.32} & 86.67 & 57.04 & 0.49  & 86.33 & 86.54 & 85.73 & 85.73 & 81.72 & 1.44  & 86.26 & 210.2 & 100.8 & 32.7 \\
    \bottomrule
    \end{tabular}%
    }
  \label{tab:cf_kt_params_time_metrics}%
  \vspace{-0.05cm}
\end{table*}%

\begin{figure*}[htbp]
\setlength{\tabcolsep}{0.5pt}
  \centering
  \resizebox{\textwidth}{!}{%
    \begin{tabular}{cc:cc}
        
        \multicolumn{2}{c}{\footnotesize TIL} & \multicolumn{2}{c}{\footnotesize DIL} \\
    
        \includegraphics[width=0.25\textwidth, trim=0.4cm 0.2cm 1.6cm 1.2cm, clip]{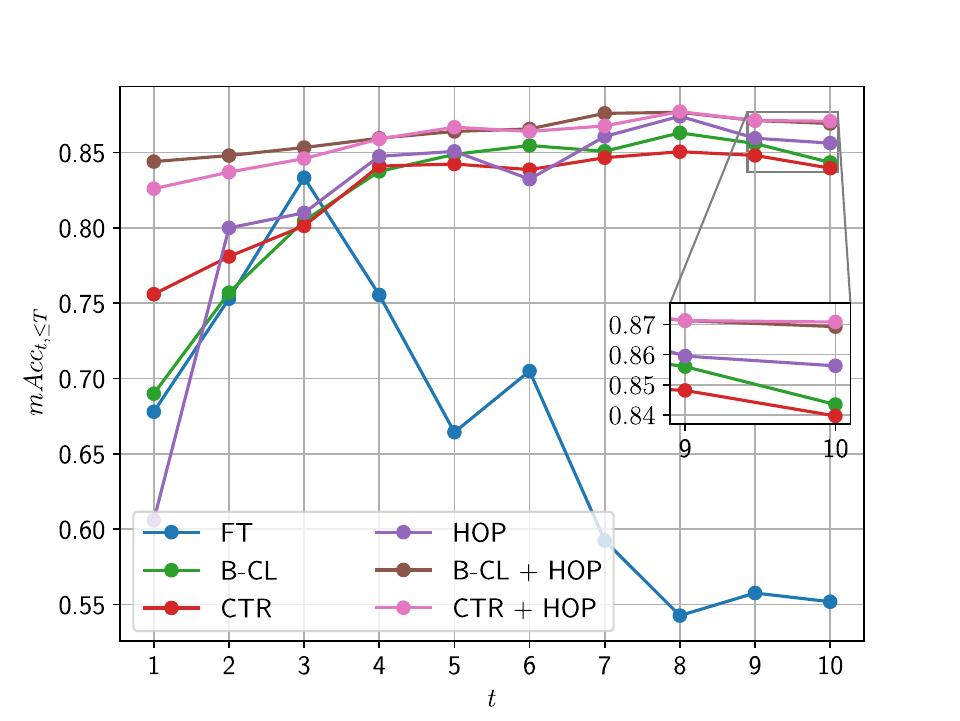} &
        \includegraphics[width=0.25\textwidth, trim=0.4cm 0.2cm 1.6cm 1.2cm, clip]{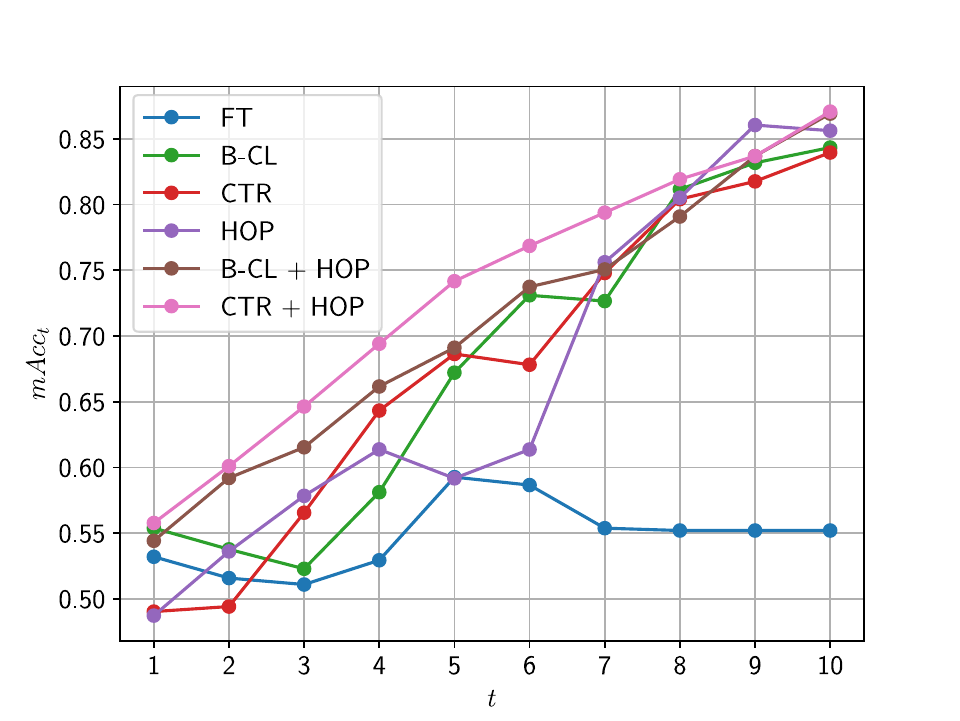} &
        
        \includegraphics[width=0.25\textwidth, trim=0.4cm 0.2cm 1.6cm 1.2cm, clip]{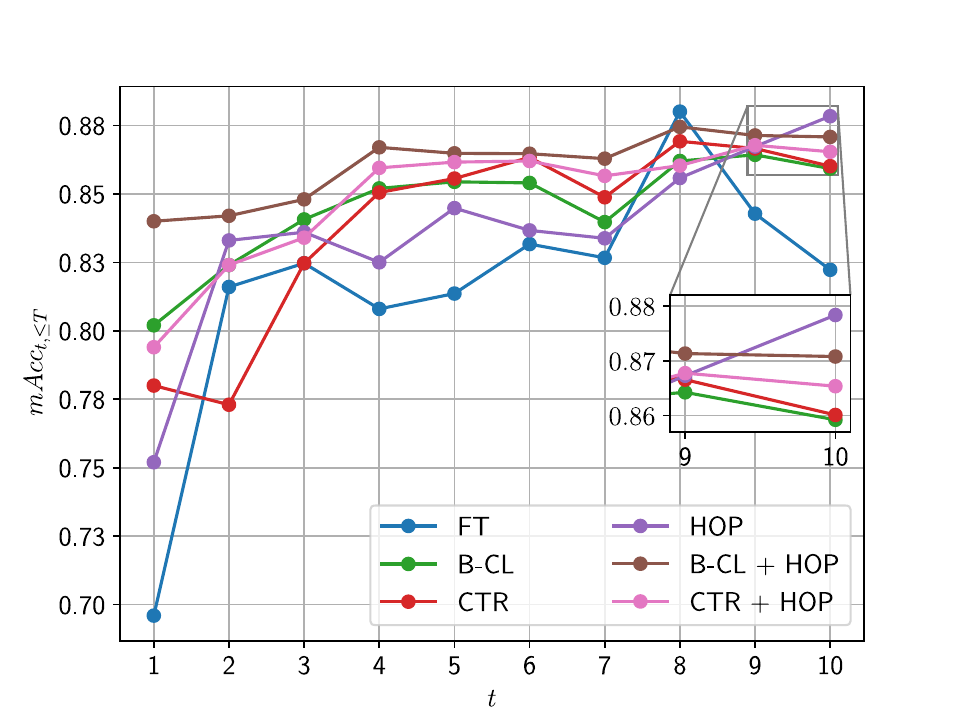} &
        \includegraphics[width=0.25\textwidth, trim=0.4cm 0.2cm 1.6cm 1.2cm, clip]{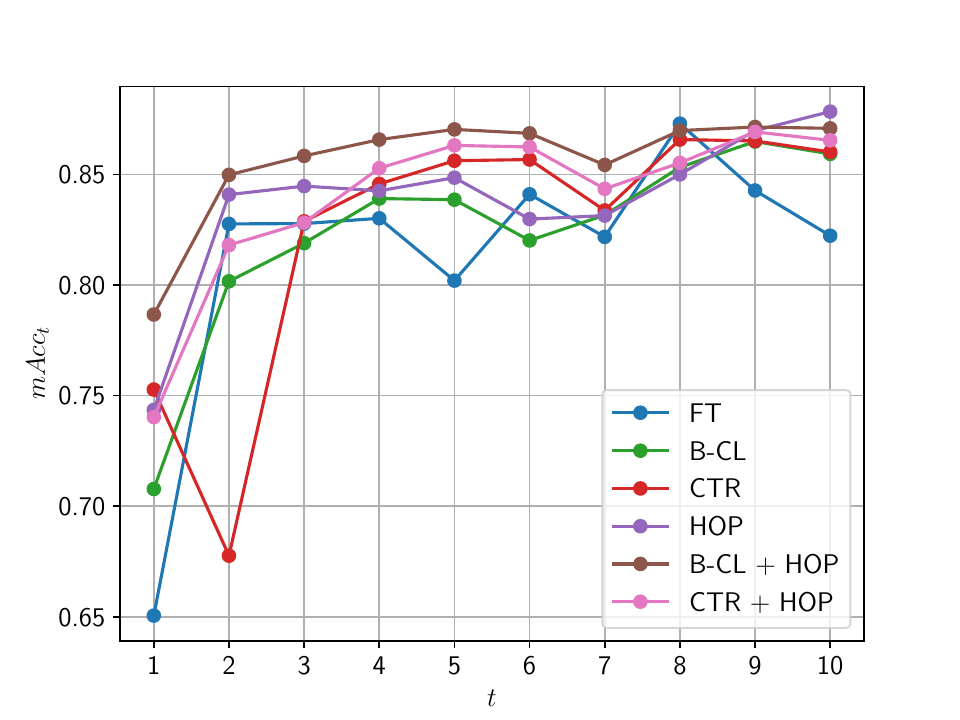} \\
    \end{tabular}%
    }
    \vspace{-0.2cm}
    \caption{Per-problem accuracy ($mAcc_t$ and $mAcc_{t,\leq T}$) on the DSC small dataset for both TIL and DIL setups.}
    \vspace{-0.3cm}
  \label{fig:per_step_accuracy}%
\end{figure*}%

\noindent\textbf{CF and KT.}
We report additional metrics to evaluate the intrinsic CF and KT properties of CL models in Table~\ref{tab:cf_kt_params_time_metrics} for both TIL and DIL in the DSC small dataset. 
Most regularization- and replay-based approaches designed for image classification (first group of eight rows) are inadequate to address CL in NLP. These methods show low accuracy due to high forgetting and low KT ($BwT$ and $FwT$), despite having good plasticity ($Pla$) to learn representations for a new problem. 
Methods designed for CL in NLP (second group of three rows), instead, can effectively increase accuracy ($mAcc$ and $MF1$) by increasing KT, reducing $Forg$ whilst maintaining $Pla$. Compared to competitors, our HOP can find a better balance between CF and KT. In both TIL and DIL, modeling high order statistics  using HOP leads to increased $mAcc$ and $MF1$ by reducing $Forg$, although showing comparable or more conservative results in terms of KT properties ($BwT$, $FwT$ and $Pla$ of HOP are not always maximized). Overall, our framework achieves a better trade-off and outperforms methods proposed specifically for TIL and for DIL.

\begin{table}[t]
\setlength{\tabcolsep}{1.1pt}
\footnotesize
  \centering
  \caption{Combination of HOP with competing CL methods on the DSC small dataset.}
  \resizebox{\linewidth}{!}{%
    \begin{tabular}{cl:cc:cc:c}
    \toprule
          & \multirow{2}[0]{*}{\textbf{HOP +}}      & \multicolumn{2}{c}{\textbf{TIL}} & \multicolumn{2}{c}{\textbf{DIL}} &  \\\cmidrule(lr){3-4} \cmidrule(lr){5-6}
          &       & $\mathbf{mAcc}$  & $\mathbf{MF1}$   & $\mathbf{mAcc}$  & $\mathbf{MF1}$   & \textbf{Time [min]} \\
          \midrule
            & FT & 85.63 \color{OliveGreen}\color{OliveGreen}\scriptsize(+55.2\%) & 84.18 \color{OliveGreen}\scriptsize(+138.6\%) & \textbf{87.94} \color{OliveGreen}\scriptsize(+7.0\%) & \textbf{87.04} \color{OliveGreen}\scriptsize(+7.0\%) & 1.3 \scriptsize(+8.3\%) \\
          & L2  & 82.57 \color{OliveGreen}\scriptsize(+16.5\%) & 80.02 \color{OliveGreen}\scriptsize(+15.8\%) & 85.60 \color{OliveGreen}\scriptsize(+2.2\%) & 84.84 \color{OliveGreen}\scriptsize(+2.4\%) & 1.4  \scriptsize(+0.0\%) \\
          & A-GEM  & 86.87  \color{OliveGreen}\scriptsize(+46.4\%) & 85.96 \color{OliveGreen}\scriptsize(+58.6\%) & 86.91 \color{OliveGreen}\scriptsize(+0.8 \%) & 86.08 \color{OliveGreen}\scriptsize(+0.7\%) & 8.8  \scriptsize(+1.1\%) \\
          & DER++ & 85.49 \color{OliveGreen}\scriptsize(+35.5\%) & 83.97 \color{OliveGreen}\scriptsize(+35.5\%) & 87.00 \color{OliveGreen}\scriptsize(+3.6\%) & 86.05 \color{OliveGreen}\scriptsize(+3.5\%) & 1.5  \scriptsize(+25.0\%) \\
          & EWC & 84.43 \color{OliveGreen}\scriptsize(+44.7\%) & 83.30 \color{OliveGreen}\scriptsize(+94.4\%) & 86.91 \color{BrickRed}\scriptsize(-0.4\%) & 86.14 \color{BrickRed}\scriptsize(-0.2 \%) & 1.4  \scriptsize(+7.7\%) \\
          & B-CL & 86.93 \color{OliveGreen}\scriptsize(+3.1\%) & 86.25 \color{OliveGreen}\scriptsize(+3.8\%) & 87.08 \color{OliveGreen}\scriptsize(+1.4\%) & 86.30 \color{OliveGreen}\scriptsize(+1.4\%) & 4.9  \scriptsize(+2.1\%) \\
          & CTR & \textbf{87.08} \color{OliveGreen}\scriptsize(+3.7\%) & \textbf{86.32} \color{OliveGreen}\scriptsize(+4.0\%) & 86.54 \color{OliveGreen}\scriptsize(+0.6\%) & 85.73 \color{OliveGreen}\scriptsize(+0.7\%) & 32.7  \scriptsize(+6.2\%) \\
          \bottomrule
    \end{tabular}%
    }
  \label{tab:HOP_plus_others}%
\end{table}%

\noindent\textbf{HOP Improves Other CL Methods.} To ensure that HOP is beneficial to CL in NLP applications, we include it in competing CL methods and report the results in Table~\ref{tab:HOP_plus_others} for DSC small in both TIL and DIL. Comparing the results against Table~\ref{tab:main_results} (gains are reported within brackets in Table~\ref{tab:HOP_plus_others} for convenience) emerges clear how HOP improves CL methods almost every time, only exception for EWC DIL. In some cases, we observe a large gain up to about $140\%$.
The gain is experienced in both TIL and DIL, the former being more largely improved by our HOP.
The integrated methods robustly outperform the original methods along all the evaluated metrics.
Remarkably, also current state-of-the-art approaches as B-CL and CTR are significantly improved by our framework.

\noindent\textbf{Per-Problem Acc} is shown in Figure~\ref{fig:per_step_accuracy}.
Acc evolution over problem is measured by $mAcc_t$ (\ie, per-problem accuracy averaged over all problems) and $mAcc_{t,\leq T}$ (\ie, per-problem accuracy averaged over the problems seen so far). FT exhibits a clear performance drop due to CF and inability to perform KT.
Methods designed for CL in NLP show an almost perfect monotonically increasing behaviour of $mAcc_{t,\leq T}$, since they are capable of learning new problems (high plasticity) without forgetting previous ones.

\noindent\textbf{Efficiency.}
HOP only adds a small increase in parameters and computation time.
We observe in Table~\ref{tab:cf_kt_params_time_metrics} that adapters of Adapter-BERT account only for about $40\%$ (73.8M) of the total number of parameters (183.3M), while clearly outperforming architectures with only a linear or convolution-based trainable head (see Table~\ref{tab:main_results}).
Compared to FT, HOP introduces just about $3\%$ more total parameters, increasing the average running training time per problem by about $8\%$ ($1.2$ to $1.3$ min).
We confirm in Table~\ref{tab:HOP_plus_others} that HOP only adds a minimal increase in computation time when added on top of existing CL methods. On average, HOP increases the mean running time per problem by just $7.2\%$.

\begin{figure}[t]
\centering
    \includegraphics[trim=0.1cm 0cm 1cm 1cm, clip, width=0.95\linewidth]{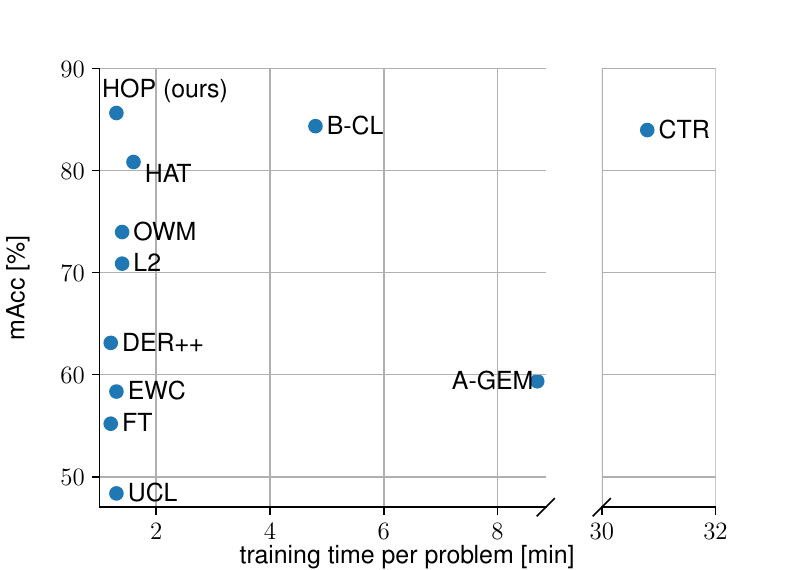}
    \caption{mAcc vs.\ training time per problem on the TIL setup. Optimal results are in the top-left corner.}
    \vspace{-0.5cm}
    \label{fig:acc_vs_time}
\end{figure}

We further highlight the effectiveness of HOP in Figure~\ref{fig:acc_vs_time}. Our approach is more computationally efficient than existing methods while outperforming them in terms of accuracy.
In particular, HOP is much faster than the main competitors being about $24\times$ faster than CTR, $4\times$ faster than B-CL. %

\begin{table}[t]
  \scriptsize
  \setlength{\tabcolsep}{2pt}
  \centering
  \caption{Ablation results on DSC small for different pooling schemes and HOP with different values of $R$.}
    \begin{tabular}{lcccc}
    \toprule
          & \multicolumn{2}{c}{\textbf{TIL}} & \multicolumn{2}{c}{\textbf{DIL}} \\\cmidrule(lr){2-3}\cmidrule(lr){4-5}
          & $\mathbf{mAcc}$  & $\mathbf{MF1}$   & $\mathbf{mAcc}$  & $\mathbf{MF1}$ \\\midrule
    \texttt{[CLS]} \cite{delvin2019bert} & 55.19 & 35.28 & 82.22 & 81.32 \\
    AVG \cite{lecun1998gradient}   & 81.44 & 80.53 & 82.94 & 81.68 \\
    MAX \cite{riesenhuber1999hierarchical}  & 79.49 & 78.48 & 82.63 & 81.26 \\
    AVGMAX \cite{monteiro2020performance} & 81.51 & 80.47 & 83.34 & 82.27 \\
    TSDP \cite{wang2021revisiting} & 77.52 & 76.30 & 78.91 & 77.50 \\
    iSQRT-COV \cite{li2018towards} & 81.47 & 80.59 & 79.62 & 78.06 \\
    \hdashline
    HOP $p=2$ (ours) & 83.52 & 82.53 & 86.91 & 86.12 \\
    HOP $p=3$ (ours) & \textbf{85.63} & 84.18 & \textbf{87.84} & \textbf{86.96} \\
    HOP $p=4$ (ours) & 84.47 & 83.58 & 86.65 & 85.81 \\\hdashline
    HOP with $m_1=$\ \texttt{[CLS]} (ours) & 85.47 & \textbf{84.61} & 87.50 & 86.70 \\
    \bottomrule
    \end{tabular}%
  \label{tab:other_pooling}%
  \vspace{-0.5cm}
\end{table}%

\noindent\textbf{Other Pooling Schemes and Order of HOP.}
Next, we observe in Table~\ref{tab:other_pooling} how popular pooling schemes underperform our solution.
As a baseline, \texttt{[CLS]} token is used for the final classification \cite{delvin2019bert}: its low results are due to the variable distribution of tokens over training.
AVG pooling \cite{lecun1998gradient} already shows remarkable improvements especially in handling the problem variability in TIL.
MAX pooling \cite{riesenhuber1999hierarchical} has a slightly worse effect than AVG.
Concatenating AVG and MAX (AVGMAX, \citealp{monteiro2020performance}) improves compared to using single clues alone.
Employment of second order statistics of tokens alone, \ie either standard deviation (TSDP, \citealp{wang2021revisiting}) or covariance (iSQRT-COV, \citealp{li2018towards}), improves on TIL but not on DIL compared to the baseline.
HOP with $p=2$ improves results compared to AVGMAX whilst using the same number of statistical measures from the distribution of tokens.
We observe that the best results are obtained for HOP with $p=3$.
Intuitively, features of different problems have similar distribution of first moments, while higher moments are discriminative for the specific problems. We compute the average Wasserstein distance between distributions of features of different problems to quantify this effect. The mean distance of first moments ($0.15\pm0.03$) is considerably lower than the mean distance of second ($0.36\pm0.05$) and third moments ($0.26\pm0.07$), indicating that problems are more entangled in the feature space of first moment (lower distance) than of second-third moments.  Thus, second-third moments improve accuracy. 
On the other hand, moments $>3$ show lower distance (\eg, the average distance of fourth moments is $0.04\pm0.01$) and yield lower results (Table~\ref{tab:other_pooling}). Therefore, they have been ignored.
HOP with $m_1=$\ \texttt{[CLS]} concatenates the \texttt{[CLS]} token with high order statistics and shows results similar to our framework, suggesting that the \texttt{[CLS]} token can be used in conjunction with high order statistics in par with AVG. In other words, in HOP, $m_1$ can be either AVG or \texttt{[CLS]}.

\section{Conclusion}
We proposed HOP, which, to our knowledge, is the first CL method 
for both TIL and DIL in various 
NLP applications (ASC, DSC, NLI, TC).
HOP is a novel approach to adapt a pre-trained NLP model for CL. 
HOP relies on adapter modules and auxiliary MLPs specialized for each problem.
Then, it computes high order moments of embedded tokens to extract rich sentence-wide information, opposed to relying on a single token for classification (\eg, \texttt{[CLS]}), which fails to adapt to dynamic non-stationary input distributions.
HOP encourages KT among problems and protects problem-specific knowledge reducing CF. Experiments show that HOP sets new state-of-the-art results on the most widely used CL NLP scenarios. At the same time, HOP only adds a minimal computation footprint.

\bibliography{anthology.bib,custom.bib}

\end{document}


\maketitle

In this document we include additional details regarding our framework HOP.
Namely, we discuss: 
the employed datasets in Sec.~\ref{supp:sec:datasets};
the hyperparameters and their validation search in Sec.~\ref{supp:sec:hyperparams};
further experimental analyses regarding additional competitor, MF1, plasticity and forgetting in Sec.~\ref{supp:sec:res};
finally, some implementation details are introduced in Sec.~\ref{supp:sec:implementation}.

\section{Datasets.}
\label{supp:sec:datasets}

\begin{enumerate}
    \item \textbf{Document Sentiment Classification (DSC).}
    This application is to classify each full product review into one of the two opinion classes (\textit{positive} and \textit{negative}). The training data of each problem consists of reviews of a particular type of product. The opinion polarity is generally predicted on top of the \texttt{[CLS]} token.
    We employ a set of 10 DSC datasets (reviews of 10 products) to produce sequences of problems. The products are \textit{Sports}, \textit{Toys}, \textit{Tools}, \textit{Video}, \textit{Pet}, \textit{Musical}, \textit{Movies}, \textit{Garden}, \textit{Offices}, and \textit{Kindle} \cite{ke2020continual}. 
    We considered experiments in two scenarios: 
    \begin{itemize}
        \item using \textbf{small} data in each problem (DSC small): 100 positive and 100 negative training reviews per problem;
        \item using the \textbf{full} data in each problem (DSC full): 2500 positive and 2500 negative training reviews per problem.
    \end{itemize}
    The first experiment is arguably more useful in practice because labeling a large number of examples is very costly. The same validation reviews (250 positive and 250 negative) and the same test reviews (250 positive and 250 negative) are used in both experiments.
    
    \item \textbf{Aspect Sentiment Classification (ASC).} It classifies a review sentence on the aspect-level sentiment (one of \textit{positive}, \textit{negative}, and \textit{neutral}). For example, the sentence “The picture is great but the sound is lousy” about a TV expresses a positive opinion about the aspect “picture” and a negative opinion about the aspect “sound”. 
    We followed the same setup as \cite{xu2019bert}, where the aspect term and sentence are concatenated via \texttt{[SEP]} token and the opinion is generally predicted on top of the \texttt{[CLS]} token.
    We employ a set of 19 ASC datasets (review sentences of 19 products) to produce sequences of problems. Each dataset represents a problem. The datasets are from 4 sources: (1) HL5Domains \cite{hu2004mining} with reviews of 5 products; (2) Liu3Domains \cite{liu2015automated} with reviews of 3 products; (3) Ding9Domains \cite{ding2008holistic} with reviews of 9 products; and (4) SemEval14 with reviews of 2 products - SemEval 2014 Task 4 \cite{pontiki2014semeval} for laptop and restaurant. For (1), (2) and (3), we use about 10\% of the original data as the validation data, another about 10\% of the original data as the testing data. For (4), we use 150 examples from the training set for validation. To be consistent with existing research \cite{ke2021achieving,ke2021classic}, sentences with conflict opinions about a single aspect are not used. More detailed statistics of the 19 datasets are given in \cite{ke2021achieving}.

    \item \textbf{20News.} This dataset \cite{lang1995newsweeder} has 20 classes and each class has about 1000 documents, which we use for Text Classification (TC). The data split for train/validation/test is 1600/200/200. We created 10 problems with 2 classes per problem (in DIL, samples are revealed incrementally, however the number of classes is known \textit{a priori}). Since this is topic-based text classification data, the classes are very different and have little shared knowledge. This application (and dataset) is mainly used to show CL models' ability to avoid forgetting.
     
    \item \textbf{Natural Language Inference (NLI).} We target NLI for sentence understanding using the MultiNLI dataset \cite{williams2018broad}, one of the largest corpus of its kind. Sentences are classified into 3 classes: \textit{entailment}, \textit{neutral} and \textit{contradiction}. We divide data in 5 problems, each belonging to a specific domain (\textit{fiction}, \textit{telephone}, \textit{slate}, \textit{travel}, and \textit{government}) \cite{asghar2019progressive}. The data split for train/validation/test is 353431/39271/9815 divided across the 5 problems.

\end{enumerate}

\begin{figure*}[t]
\setlength{\tabcolsep}{0.5pt}
  \centering
  \resizebox{\linewidth}{!}{%
    \begin{tabular}{cc:cc}
        \multicolumn{2}{c}{\footnotesize TIL} & \multicolumn{2}{c}{\footnotesize DIL} \\
    
        \includegraphics[width=0.25\textwidth, trim=0.4cm 0.2cm 1.6cm 1.2cm, clip]{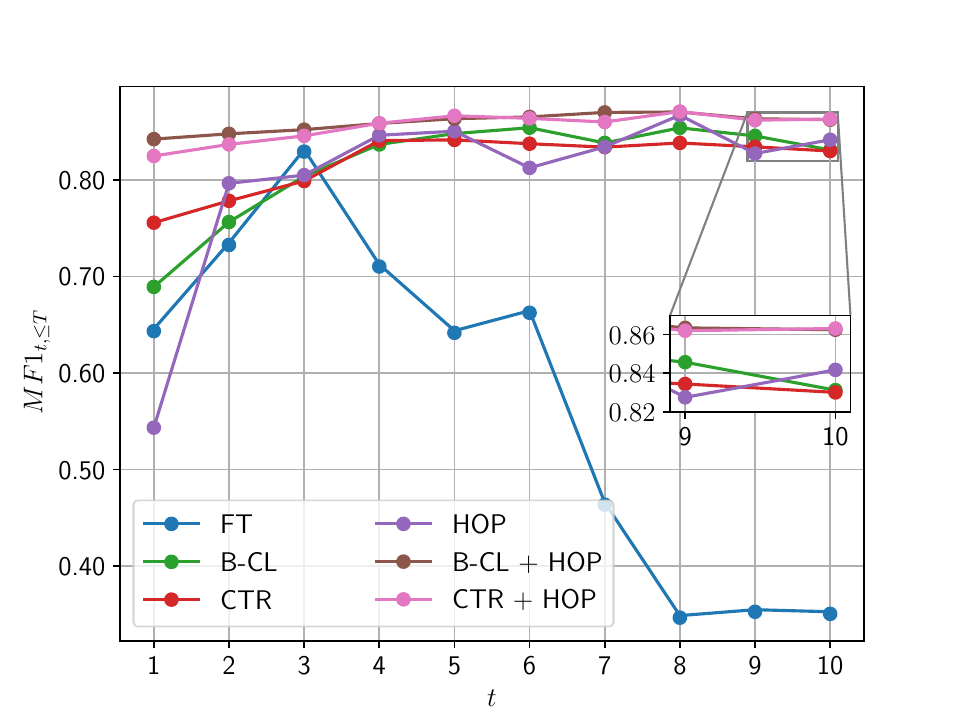} &
        \includegraphics[width=0.25\textwidth, trim=0.4cm 0.2cm 1.6cm 1.2cm, clip]{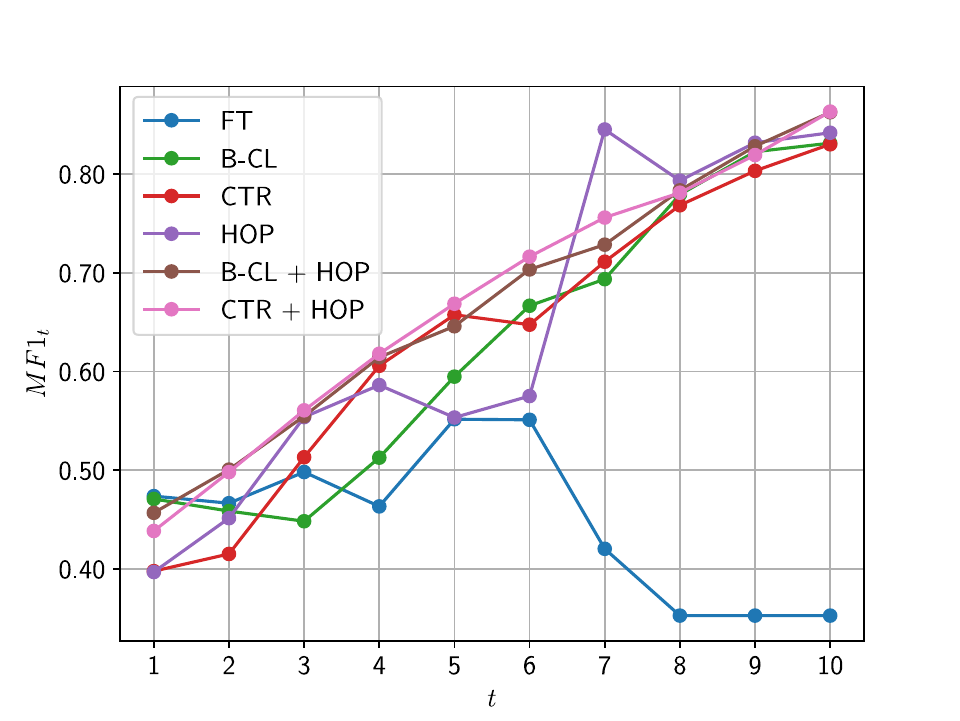} &
        
        \includegraphics[width=0.25\textwidth, trim=0.4cm 0.2cm 1.6cm 1.2cm, clip]{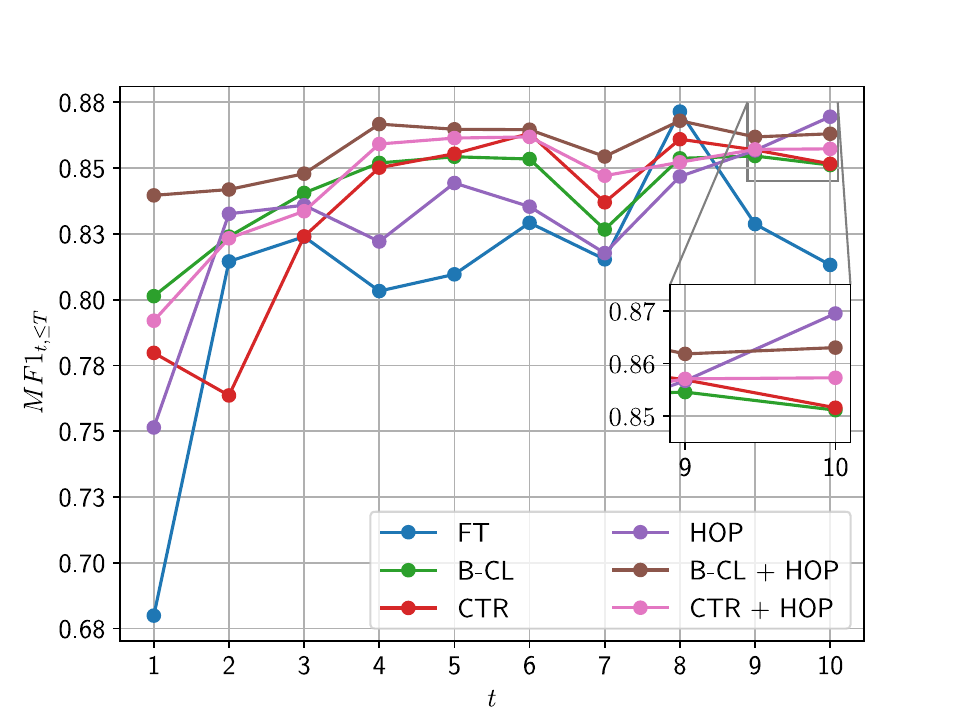} &
        \includegraphics[width=0.25\textwidth, trim=0.4cm 0.2cm 1.6cm 1.2cm, clip]{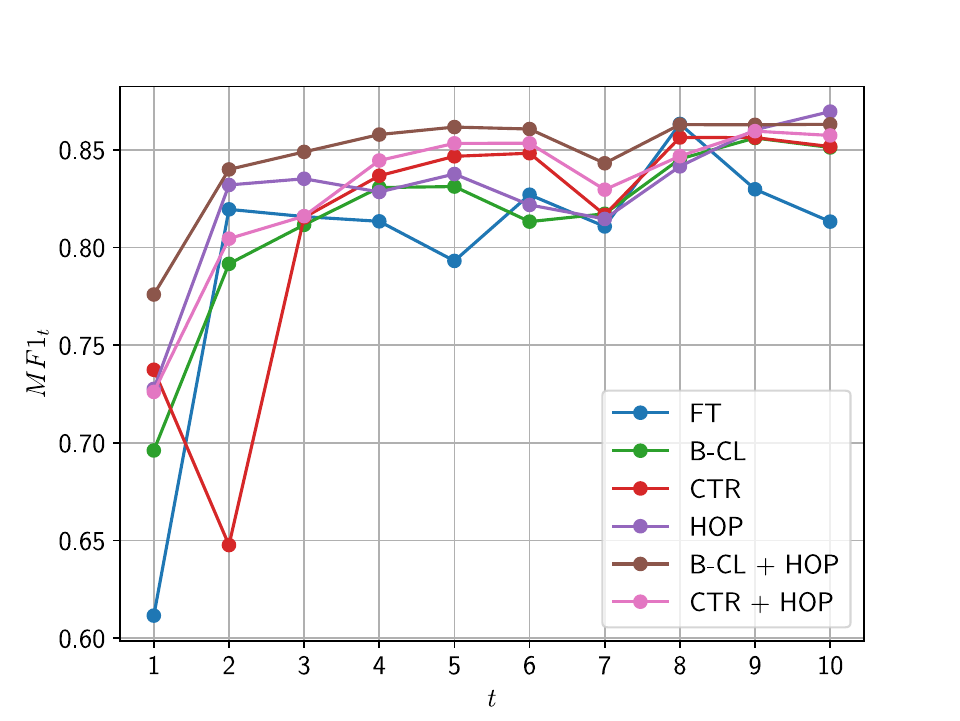} \\
    \end{tabular}%
    }
    \caption{Per-problem $MF1$ ($MF1_t$ and $MF1_{t,\leq T}$) on the DSC small dataset for both TIL and DIL setups.}
  \label{fig:per_step_MF1}%
\end{figure*}%

\section{Hyperparameters and Validation Search.}
\label{supp:sec:hyperparams}

In the main paper we give the values for the main hyperparameters of our work; here we further elucidate how they have been selected. The results reported in the paper are given on the test set in correspondence of the parameters with the highest validation performance.

Regarding hyperparameter search (selected value in \textbf{bold}), we performed grid search on the  batch size within $\{16, \mathbf{32}, 64\}$, learning rate over ${\{ 10^{-6}, 2\cdot 10^{-5}, \mathbf{5\cdot 10^{-5}}, 10^{-4} \}}$.
We use Adam optimizer with learning rate equal to $5\cdot 10^{-5}$ and batch size of $32$.
We use $20$ epochs for DSC and $10$ epochs for all other datasets; early-stopping is enabled with patience of $5$ epochs computed on the validation loss during training for each problem.

The only hyperparameter specific to our framework is $p$, which is set according to the best validation performance from a grid search over $\{1,2,\mathbf{3},4,5\}$. Empirically $p=3$ is found to provide the best results and represents a good compromise in terms of additional computational complexity.
The maximum total input sequence length is set to $128$ which is long enough for all the considered datasets.

We adopt BERT$_{\mathrm{BASE}}$ (uncased) as our unreduced tokenizer $\mathcal{T}'$, whose training follows \citet{xu2019bert}. For a given input sequence $x_t^k$, $\mathcal{T}'$ produces tokens ${h_{t,d}^k}$ where $d$ spans the dimensionality of the embedded sequence and each $h_{t,d}^k \in \mathbb{R}^{Q}$, where $Q=768$ is the channel size.
The personalized MLP heads have a softmax output used for classification in the last layer of the BERT, and they are trained with the categorical cross-entropy loss. 
Following previous works \cite{ke2021achieving}, we enable dropout on the extracted features: rate is set to $0.5$ on top of Bert (Frozen) + CNN, while it is set to $0.1$ between fully-connected layers of Adapter-BERT.

\section{Additional Experimental Results.}
\label{supp:sec:res}

\noindent\textbf{Restricted Comparison against LAMOL.}
We compare our method against several competing approaches that can be applied to a BERT architecture.
Therefore, in the main paper we did not include LAMOL \cite{sun2019lamol} in our comparisons, because it uses the large language model GPT-2 \cite{radford2019language}. However, a restricted comparison is reported in Tab.~\ref{tab:lamol} in four benchmarks in the TIL setup. We observe that no clear winner exists, with LAMOL outperforming HOP in DSC, while being outperformed by our HOP in both ASC and TC.
Once again, we remark that LAMOL is based on GPT-2 and cannot be easily extended to BERT-based architectures. Indeed, LAMOL requires models with high generative power in order to generate pseudo-samples of past problems. 
Hence, it is not directly comparable with our HOP and other concurrent approaches on CL for NLP.

\begin{table*}[htbp]
  \centering
  \caption{Reduced evaluation against LAMOL. *: results extracted from \cite{ke2021achieving}.}
    \begin{tabular}{lcccccccc}
    \toprule
          \multicolumn{1}{c}{\multirow{2}[0]{*}{\textbf{CL Method}}} & \multicolumn{2}{c}{\textbf{ASC}} & \multicolumn{2}{c}{\textbf{DSC small}} & \multicolumn{2}{c}{\textbf{DSC full}} & \multicolumn{2}{c}{\textbf{20News}} \\
          \cmidrule(lr){2-3}\cmidrule(lr){4-5}\cmidrule(lr){6-7}\cmidrule(lr){8-9}
          & \textbf{mAcc} & \textbf{MF1} & \textbf{mAcc} & \textbf{MF1} & \textbf{mAcc} & \textbf{MF1} & \textbf{mAcc} & \textbf{MF1} \\\midrule
    CTR (Adapter-BERT)   & 89.47 & 83.62 & 83.96 & 83.00 & 89.31 & 88.75 & 95.25 & 95.23 \\
    LAMOL* (GPT-2) & 88.91 & 80.59 & \textbf{89.12} & \textbf{86.58} & \textbf{92.11} & \textbf{91.72} & 66.13 & 75.74 \\
    HOP (Adapter-BERT)   & \textbf{89.84} & \textbf{85.06} & 85.63 & 84.18 & 90.08 & 89.44 & \textbf{95.30} & \textbf{95.29} \\
    \bottomrule
    \end{tabular}%
  \label{tab:lamol}%
\end{table*}%

\begin{figure}[t]
\centering
    \includegraphics[trim=0.1cm 0cm 1cm 1cm, clip, width=1\linewidth]{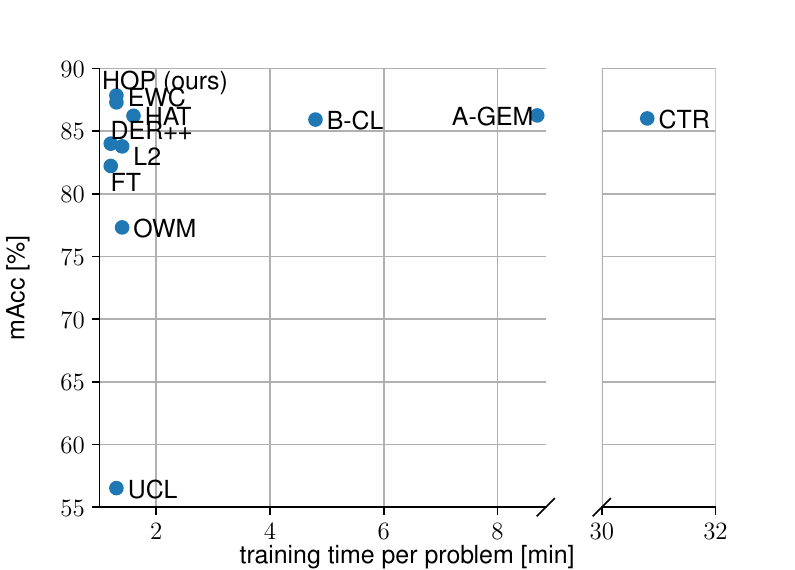}
    \vspace{-0.4cm}
    \caption{mAcc versus training time per problem of the compared approaches on the DIL setup. Optimal results are in the top-left corner.}
    \vspace{-0.5cm}
    \label{fig:acc_vs_time_DIL}
\end{figure}

\noindent\textbf{Efficiency} plot on the DIL setup is reported in Fig.~\ref{fig:acc_vs_time_DIL}.
Similarly to the main plot reported in the main paper, we observe that our HOP is significantly more efficient than the compared approaches, while outperforming them in terms of average accuracy.

\noindent\textbf{Per-Problem MF1.} We report in Fig.~\ref{fig:per_step_MF1} the evolution over problem indices of $MF1_t$ (\ie, per-problem $MF1$ averaged on all problems) and $MF1_{t,\leq T}$ (\ie, per-problem $MF1$ averaged over the problems seen so far), similarly to the results presented in the main paper for $mAcc_t$ and $mAcc_{t,\leq T}$. FT exhibits a clear performance drop due to CF and inability to perform KT.
Methods designed for CL in NLP show an almost perfect monotonically increasing behaviour of $MF1_{t,\leq T}$ since they are capable of learning new problems (high plasticity) without forgetting previous ones.

\begin{figure}[t]
\setlength{\tabcolsep}{0.5pt}
  \centering
  \resizebox{\linewidth}{!}{%
    \begin{tabular}{cc}
       \footnotesize TIL & \footnotesize DIL \\
    
        \includegraphics[width=0.25\textwidth, trim=0.8cm 0.2cm 1.6cm 1.2cm, clip]{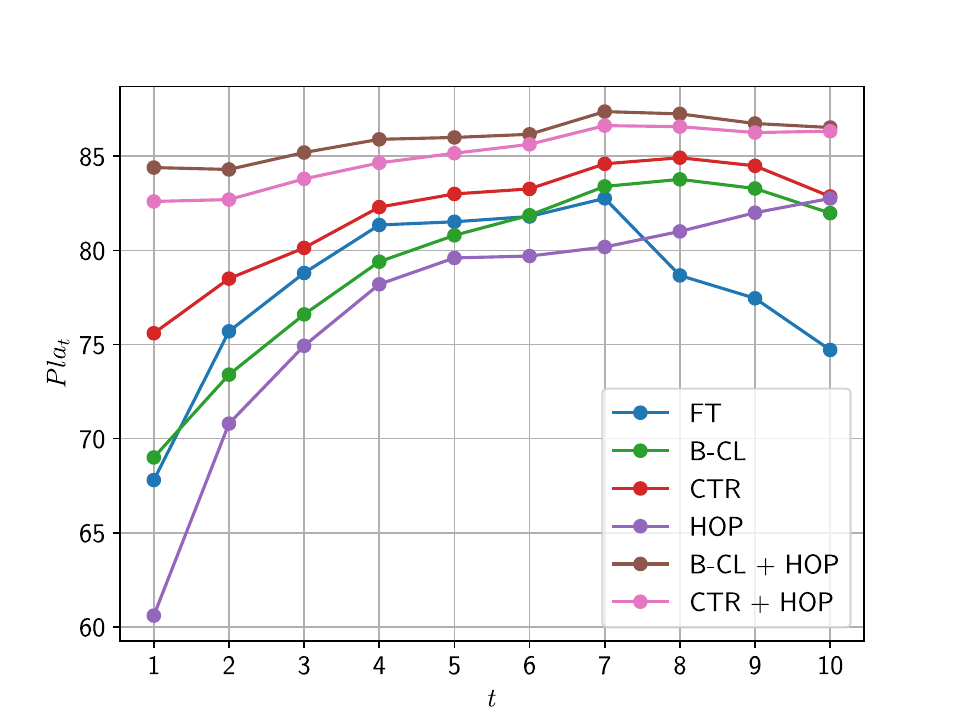} &
        \includegraphics[width=0.25\textwidth, trim=0.8cm 0.2cm 1.6cm 1.2cm, clip]{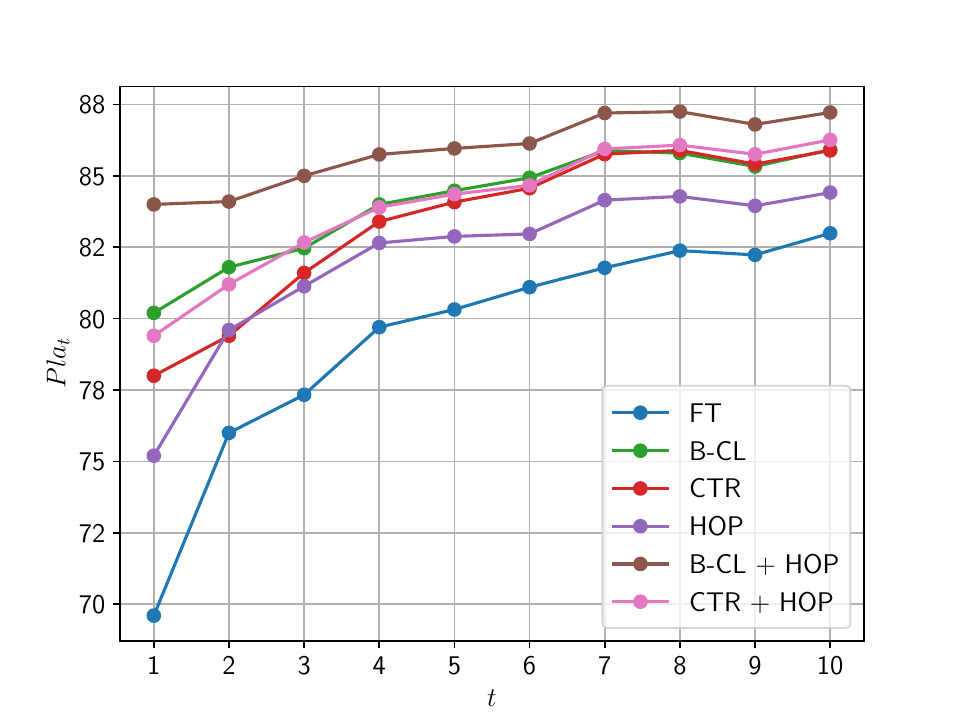} \\
    
    \end{tabular}%
    }
    \caption{Per-problem plasticity, $Pla_t$, on the DSC small dataset for both TIL and DIL setups.}
  \label{fig:per_step_plasticity}%
\end{figure}%

\noindent\textbf{Per-Problem Plasticity} is shown in Fig.~\ref{fig:per_step_plasticity}.
Compared to existing baselines, we observe that HOP is more conservative in terms of plasticity and adaptability to the new problems, especially in the TIL setup. This is due to the implicit regularization effect brought in by the high order pooling, which extracts overall tendency while reducing point-wise problem accuracy.
In other words, a sub-optimal plasticity to the current problem $t$ is the price we pay to maintain higher accuracy over all problems in the stability-plasticity trade-off dilemma \cite{mccloskey1989catastrophic}.

On the other hand, combining HOP with B-CL \cite{ke2021adapting} and CTR \cite{ke2021achieving}, we experience a high increase in plasticity in both TIL and DIL setups.

\begin{figure}[t]
\setlength{\tabcolsep}{0.5pt}
  \centering
  \resizebox{\linewidth}{!}{%
    \begin{tabular}{cc}
       \footnotesize TIL & \footnotesize DIL \\
    
        \includegraphics[width=0.25\textwidth, trim=0.8cm 0.2cm 1.6cm 1.2cm, clip]{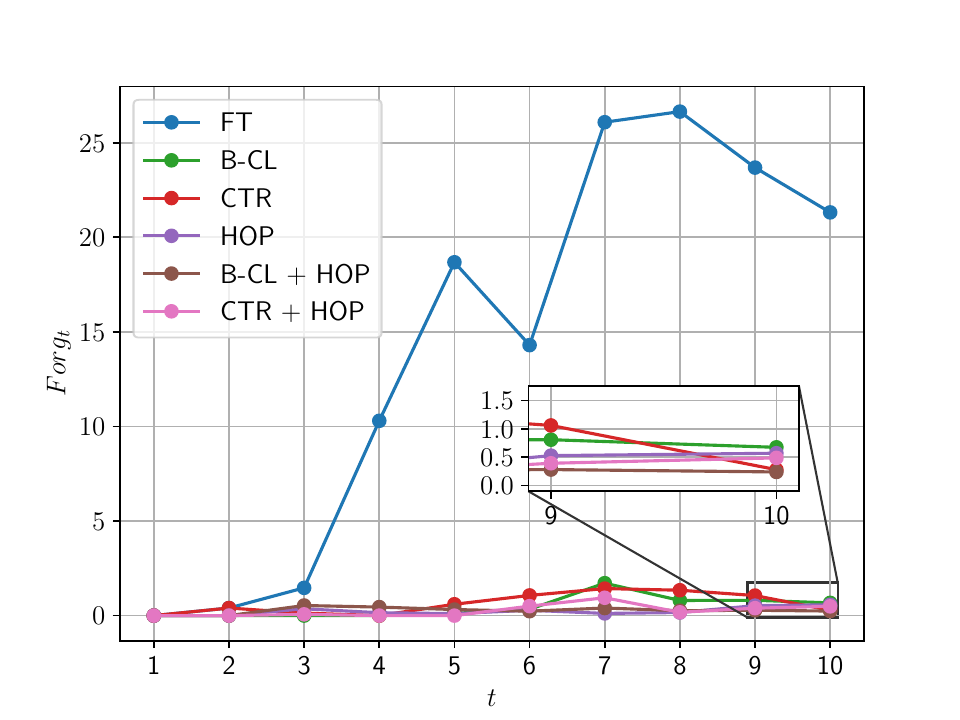} &
        \includegraphics[width=0.25\textwidth, trim=0.8cm 0.2cm 1.6cm 1.2cm, clip]{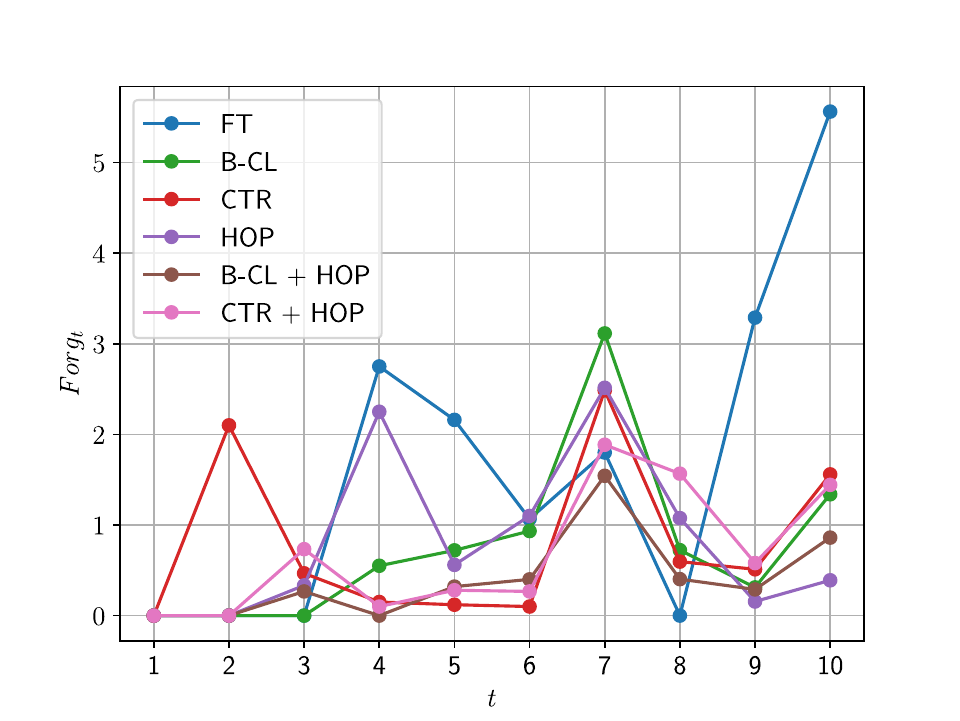} \\
    
    \end{tabular}%
    }
    \caption{Per-problem forgetting, $Forg_t$, on the DSC small dataset for both TIL and DIL setups.}
  \label{fig:per_step_forgetting}%
\end{figure}%

\noindent\textbf{Per-Problem Forgetting} is reported in Fig.~\ref{fig:per_step_forgetting}. FT generally achieves the worst results, having no mechanisms to avoid CF.
Our method can effectively reduce forgetting thanks to the high order pooling that ease the discrimination task between sentences of different problems. Therefore, HOP is among the best methods in both TIL and DIL, either used alone or in conjunction with competing approaches.

\section{Implementation Details.}
\label{supp:sec:implementation}
All CL baselines are implemented in PyTorch on a single codebase to guarantee fair comparisons. 
For all CL baselines, we use the code provided by the authors and adapted for our NLP applications using their original parameters. %
Our codebase is extended from \url{https://github.com/ZixuanKe/PyContinual} to include our framework and all the considered scenarios. The results are evaluated on a single NVIDIA GeForce RTX 3090 GPU supported by an Intel(R) Xeon(R) Gold 5218R CPU @ 2.10GHz with 80 cores and 252GB RAM.

\bibliography{anthology.bib,custom.bib}